\documentclass[10pt,journal,compsoc]{IEEEtran}
\ifCLASSOPTIONcompsoc
  \usepackage[nocompress]{cite}
\else
  \usepackage{cite}
\fi
\ifCLASSINFOpdf
\else
\fi
\ifCLASSOPTIONcompsoc
	\usepackage[caption=false,font=footnotesize,labelfont=sf,textfont=sf]{subfig}
\else
  \usepackage[caption=false,font=footnotesize]{subfig}
\fi

\usepackage{epsfig}
\usepackage{graphicx}
\usepackage{amsmath}
\usepackage{amssymb}

\usepackage{multirow}

\usepackage{pdfrender}
\newcommand*{\boldcheckmark}{%
	\textpdfrender{
		TextRenderingMode=FillStroke,
		LineWidth=.5pt, 
	}{\checkmark}%
}

\usepackage{enumerate}
\usepackage{bm}
\usepackage{mathtools} 
\usepackage{ragged2e} 
\usepackage{enumitem}

\usepackage{subfig}
\usepackage{array}

\newcolumntype{x}[1]{>{\centering\arraybackslash}p{#1pt}}
\newcommand{\tablestyle}[2]{\setlength{\tabcolsep}{#1}\renewcommand{\arraystretch}{#2}\centering\footnotesize}
\usepackage[font=small]{caption}

\usepackage{algorithm}
\usepackage{algpseudocode}
\usepackage{footmisc}

\newcommand*\inlineimage[1]{\raisebox{-0.14\baselineskip}{\includegraphics[height=0.95\baselineskip]{#1}}}
\newcommand{\tieimg}{\inlineimage{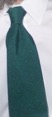}}

\newcommand{\ie}{\textit{i}.\textit{e}.}
\newcommand{\eg}{\textit{e}.\textit{g}.}
\newcommand{\aka}{\textit{a}.\textit{k}.\textit{a}.}

\newcommand{\etal}{\textit{et}.\textit{al}.}

\usepackage[pagebackref=false,breaklinks=true,colorlinks,bookmarks=false]{hyperref}
\usepackage[dvipsnames]{xcolor}
\usepackage{colortbl} 
\definecolor{mygray-bg}{gray}{0.9}

\begin{document}
%
\title{Counterfactual Samples Synthesizing and Training for Robust Visual Question Answering}
%
%
%
%

\author{Long~Chen$^*$,
	Yuhang~Zheng$^*$,
	Yulei~Niu,
	Hanwang~Zhang,
	and Jun~Xiao$^\dagger$ \\
	\IEEEcompsocitemizethanks{
		\IEEEcompsocthanksitem $^*$ These authors contributed equally, and $^\dagger$ denotes corresponding author. 
		\IEEEcompsocthanksitem Long Chen is with the Department of Computer Science and Engineering, The Hong Kong University of Science and Technology, Kowloon, Hong Kong SAR, 999077. Email: longchen@ust.hk. 
		\IEEEcompsocthanksitem Yuhang Zheng and Jun Xiao are with the College of Computer Science, Zhejiang University, Hangzhou, China, 310027. E-mails: itemzheng@zju.edu.cn, junx@zju.edu.cn. 
            \IEEEcompsocthanksitem Yulei Niu is with the Department of Electrical Engineering, Columbia University, New York, NY, USA, 10027. E-mail: yn.yuleiniu@gmail.com.		
            \IEEEcompsocthanksitem Hanwang Zhang is with the School of Computer Science and Engineering, Nanyang Technological University, Singapore, 639798. E-mails: hanwangzhang@ntu.edu.sg.
	}
}

\markboth{IEEE TRANSACTIONS ON PATTERN ANALYSIS AND MACHINE INTELLIGENCE}%
{IEEE TRANSACTIONS ON PATTERN ANALYSIS AND MACHINE INTELLIGENCE}
%



\IEEEtitleabstractindextext{%

    \justifying 

	\begin{abstract}
        Today's VQA models still tend to capture superficial linguistic correlations in the training set and fail to generalize to the test set with different QA distributions. To reduce these language biases, recent VQA works introduce an auxiliary question-only model to regularize the training of targeted VQA model, and achieve dominating performance on diagnostic benchmarks for out-of-distribution testing. However, due to the complex model design, ensemble-based methods are unable to equip themselves with two indispensable characteristics of an ideal VQA model: 1) Visual-explainable: The model should rely on the right visual regions when making decisions. 2) Question-sensitive: The model should be sensitive to the linguistic variations in questions. To this end, we propose a novel model-agnostic Counterfactual Samples Synthesizing and Training (CSST) strategy. After training with CSST, VQA models are forced to focus on all critical objects and words, which significantly improves both visual-explainable and question-sensitive abilities. Specifically, CSST is composed of two parts: \emph{Counterfactual Samples Synthesizing} (CSS) and \emph{Counterfactual Samples Training} (CST). CSS generates counterfactual samples by carefully masking critical objects in images or words in questions and assigning pseudo ground-truth answers. CST not only trains the VQA models with both complementary samples to predict respective ground-truth answers, but also urges the VQA models to further distinguish the original samples and superficially similar counterfactual ones. To facilitate the CST training, we propose two variants of supervised contrastive loss for VQA, and design an effective positive and negative sample selection mechanism based on CSS. Extensive experiments have shown the effectiveness of CSST. Particularly, by building on top of model LMH+SAR~\cite{clark2019don,si2021check}, we achieve record-breaking performance on all out-of-distribution benchmarks (\eg, VQA-CP v2, VQA-CP v1, and GQA-OOD).
	\end{abstract}
	
	\vspace{-0.5em}
	}

\maketitle


%

\IEEEraisesectionheading{\section{Introduction}\label{sec:intro}}

\begin{figure}[t]
    \centering
    \includegraphics[width=0.88\linewidth]{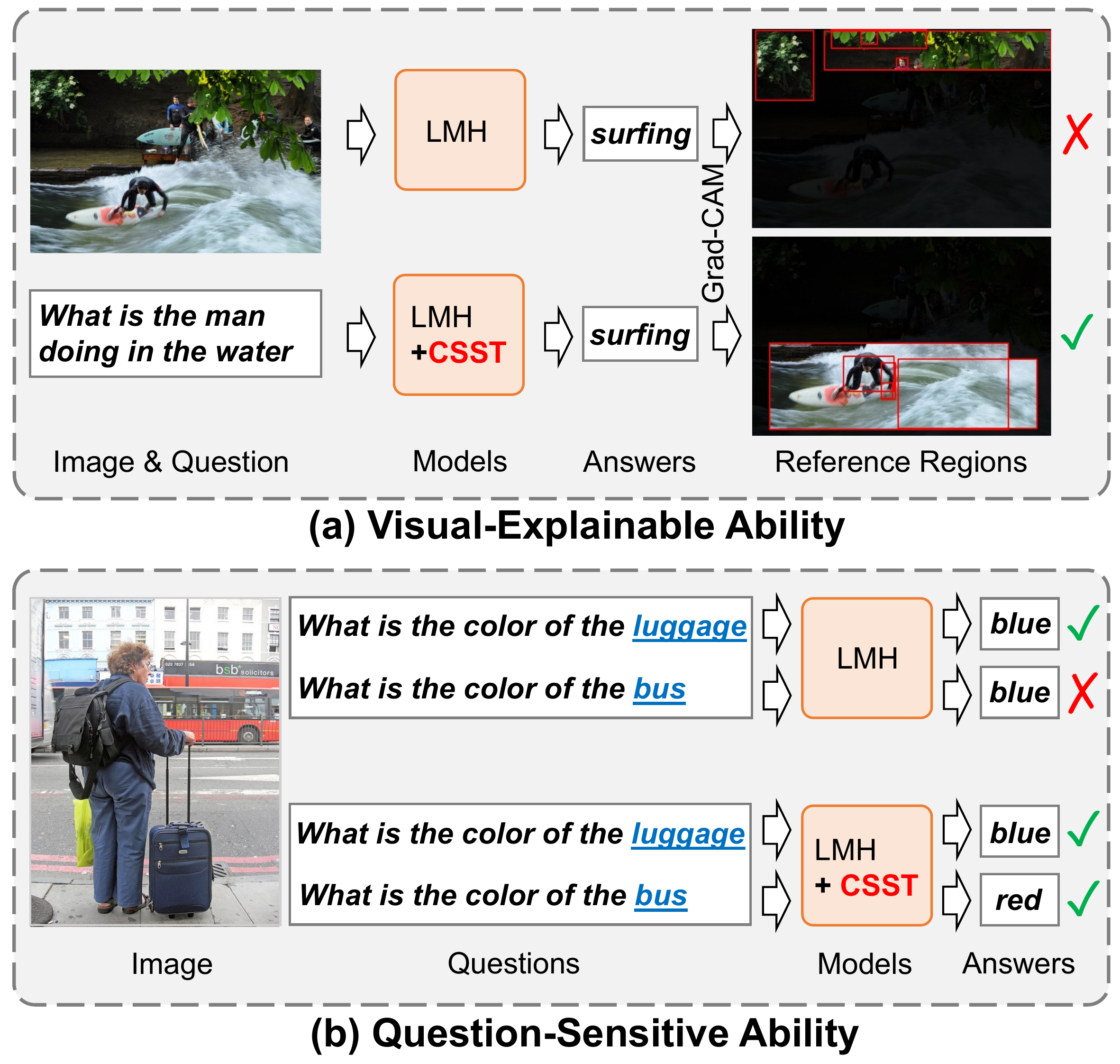}
    \vspace{-0.5em}
    \caption{Two indispensable characteristics of an ideal VQA model. (a) \textbf{Visual-explainable ability}: The model can not only predict the correct answer (\eg, ``\texttt{surfing}''), but also rely on the right visual reference regions when making this prediction. (b) \textbf{Question-sensitive ability}: The model should be sensitive to linguistic variations. For example, after replacing the critical word ``\texttt{luggage}'' with ``\texttt{bus}'', the predicted answers of the two questions should be different. LMH~\cite{clark2019don} is a SOTA VQA model.
    }
    \vspace{-1em}
    \label{fig:1}
\end{figure}

%
%
%
%


\IEEEPARstart{V}{isual} Question Answering (VQA), \ie, answering natural language questions about the given visual content, is one of the essential abilities of advanced AI agents. With the release of multiple large-scale VQA datasets, 
VQA has received unprecedented attention and hundreds of VQA models have been developed. However, since the inevitable annotation artifacts in the real image datasets, today's VQA models always over-rely on superficial linguistic correlations between the questions and answers (\aka, language biases)~\cite{agrawal2016analyzing, zhang2016yin, johnson2017clevr, goyal2017making}. For example, a model naively answering ``\texttt{2}'' for all ``\texttt{How many X?}'' questions can still get satisfactory performance regardless of ``\texttt{X}''. To disentangle the bias factors and clearly monitor the progress of VQA research, several diagnostic benchmarks have been proposed, such as VQA-CP~\cite{agrawal2018don} and GQA-OOD~\cite{kervadec2021roses}. These benchmarks deliberately keep different question-answer distributions in training and test sets. Moreover, the performance of many state-of-the-art VQA models~\cite{andreas2016neural, fukui2016multimodal, yang2016stacked, anderson2018bottom, tan2019lxmert} all consistently drop significantly on these diagnostic benchmarks compared to their counterparts (\ie, VQA v2/v1 and GQA).

\begin{figure*}[tbp]
    \centering
    \includegraphics[width=0.9\linewidth]{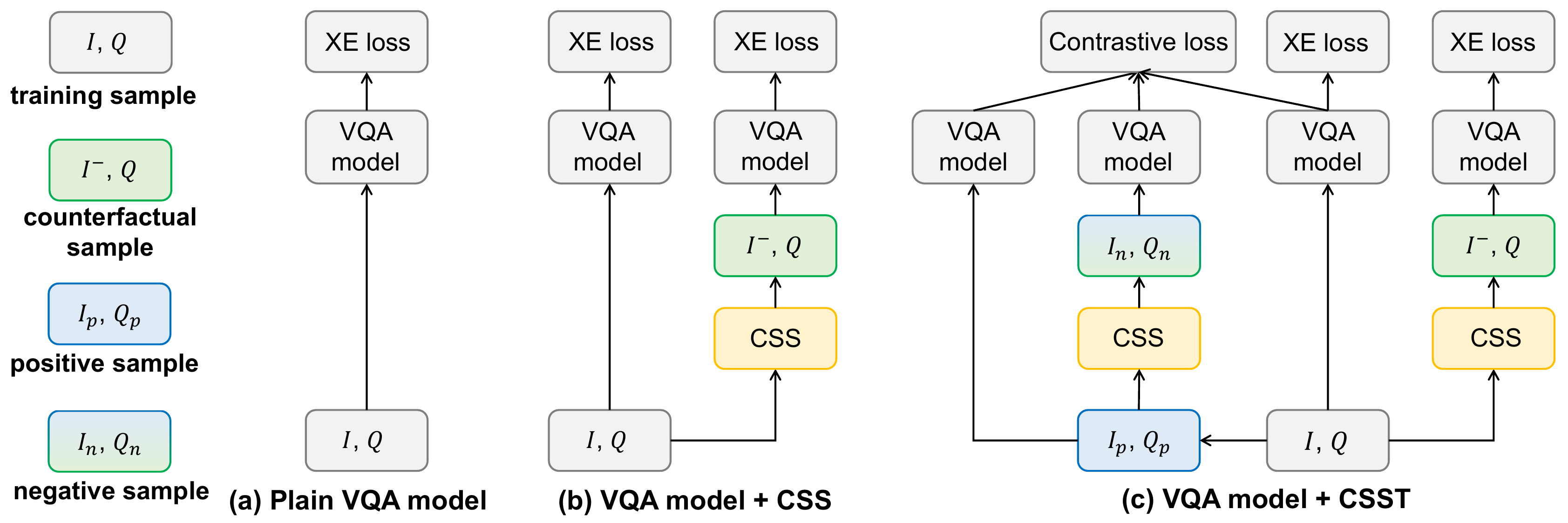}
    \vspace{-0.8em}
    \caption{Comparisons on three different VQA frameworks, and the weights of all VQA models are shared in each framework. \textbf{(a)} The most common VQA framework, \ie, a plain VQA model with XE loss using original samples. \textbf{(b)} The framework of previous CSS work~\cite{chen2020counterfactual}: Using CSS to generate counterfactual samples, and training the VQA model using XE loss for both original samples and counterfactual samples. \textbf{(c)} The framework of the proposed CSST: Besides the XE training objectives, we use CSS to further generate positive-negative sample pairs, and train the VQA model with an extra contrastive loss. For conciseness, we only show one (type of) counterfactual sample and negative sample in each figure, and we skip the answer sets for all samples in each figure.}
    \label{fig:2}
\end{figure*}

Currently, the most prevalent solutions to mitigate these language bias issues are \textbf{ensemble-based} methods: they introduce an auxiliary question-only model to regularize the training of targeted VQA model. Specifically, these methods can further be categorized into two sub-types: 1) \emph{Adversary-based models}~\cite{ramakrishnan2018overcoming, grand2019adversarial, belinkov2019don}: They train two models (\ie, the question-only and targeted VQA model) in an adversarial manner~\cite{goodfellow2014generative, chen2018zero}, \ie, minimizing the loss of VQA model while maximizing the loss of the question-only model. Since these two models are typically designed to share the same question encoder, these adversary-based methods aim to reduce language biases by learning bias-neutral question representations. Unfortunately, this adversarial training scheme brings significant noise into gradient calculations, which results in an unstable training process~\cite{grand2019adversarial}. 2) \emph{Fusion-based models}~\cite{cadene2019rubi, clark2019don, mahabadi2019simple, niu2021counterfactual, han2021greedy}: They late fuse the predicted answer distributions of the two models, and derive training gradients based on fused answer distribution. The design philosophy of these fusion-based models, is to let the targeted VQA model focuses more on samples, which can't be easily answered correctly by the question-only model.

Although the ensemble-based methods have dominated the performance on these diagnostic benchmarks, it is worth noting that current methods fail to equip themselves with two indispensable characteristics of an ideal VQA model: 1) \textbf{Visual-explainable ability}: The VQA model should rely on the right visual regions when making decisions, \ie, right for the right reasons~\cite{ross2017right}. As shown in Fig.~\ref{fig:1} (a), although both two models can predict the correct answer ``\texttt{surfing}'', they actually refer to different visual reference regions when making their respective answer predictions. 2) \textbf{Question-sensitive ability}: The VQA model should be sensitive to the linguistic variations in questions. As shown in Fig.~\ref{fig:1} (b), for two questions with a similar sentence structure (\eg,  only replacing the word ``\texttt{luggage}'' with ``\texttt{bus}''), if the meanings of the two questions are different, the model should perceive the discrepancy and make corresponding predictions.

\begin{figure}[tbp]
    \centering
    \includegraphics[width=\linewidth]{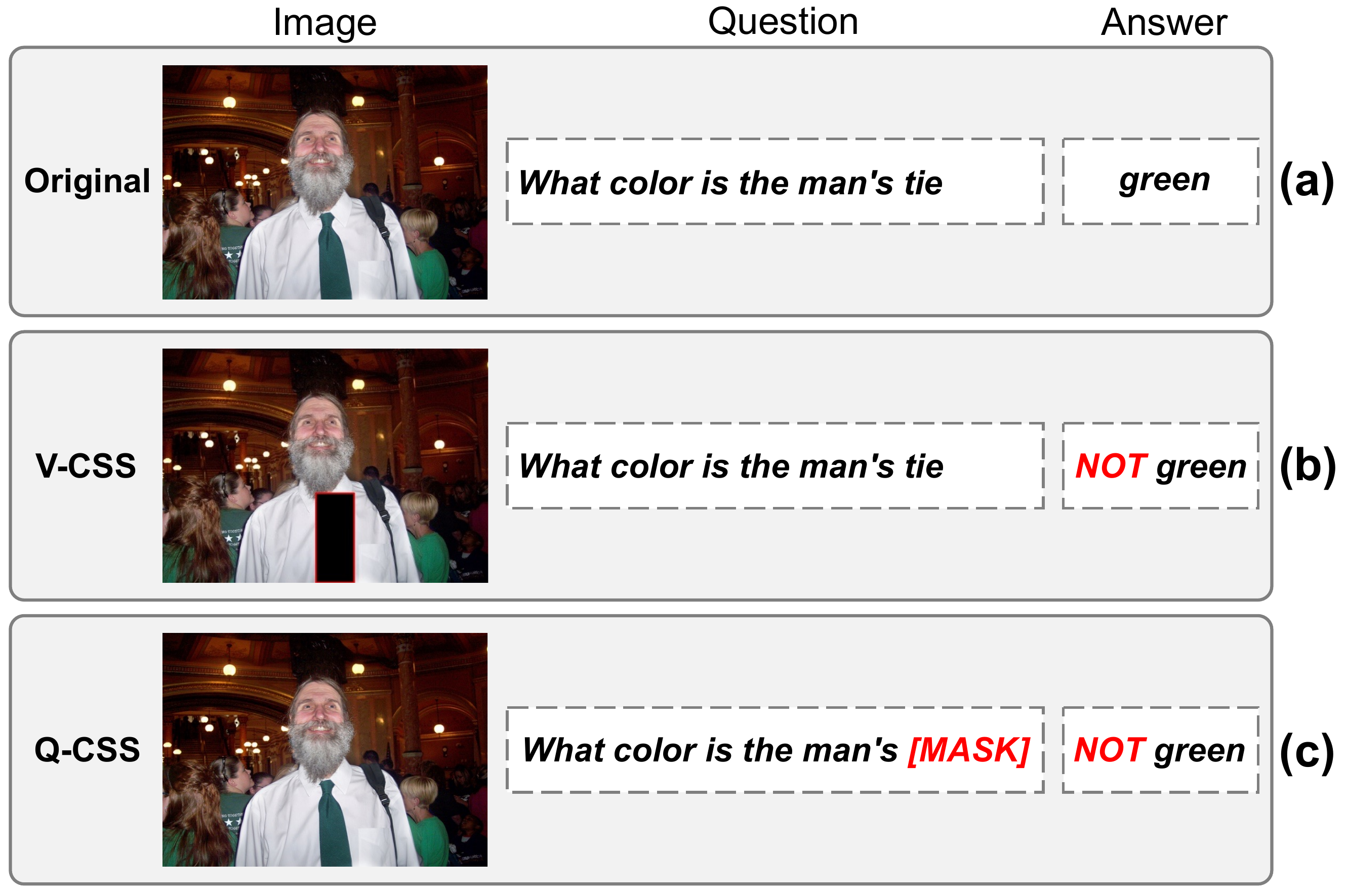}
    \vspace{-1.5em}
	\caption{(a): A training sample from VQA-CP. (b): The synthesized training sample by V-CSS. It masks critical objects in image and assigns different GT answers (\eg, ``\texttt{not green}''). (c): The synthesized training sample by Q-CSS. It replaces critical words (\eg, ``\texttt{tie}'') with special token ``[\texttt{MASK}]'' in the question and assigns different GT answers (\eg, ``\texttt{not green}'').}
    \label{fig:3}
\end{figure}

In this paper, we propose a novel model-agnostic \emph{Counterfactual Samples Synthesizing and Training} (\textbf{CSST}) strategy. CSST serves as a plug-and-play component to improve VQA models' visual-explainable and question-sensitive abilities, even for complex ensemble-based models. Specifically, CSST is composed of two parts: Counterfactual Samples Synthesizing (CSS) and Counterfactual Samples Training (CST).

\noindent\textbf{CSS.} For each original VQA training sample, CSS can generate a corresponding counterfactual sample. As shown in Fig.~\ref{fig:3}, CSS consists of two  different types of sample synthesizing mechanisms: \emph{V-CSS} and \emph{Q-CSS}. For V-CSS, it synthesizes counterfactual images by masking critical objects in the original image. By ``critical'', we mean that these objects are important for answering a certain question. For example, object~\tieimg~(\ie, green tie) is a critical object for question ``\texttt{What color is the man's tie?}''. Then, the counterfactual image and original question compose a new image-question (VQ) pair. For Q-CSS, it synthesizes counterfactual questions by replacing critical words in original question with a special token ``[\texttt{MASK}]''. Similarly, the counterfactual question and original image compose a new VQ pair. Meanwhile, to avoid expensive manual annotations, we design a dynamic answer assigning mechanism to approximate ground-truth answers for all synthesized VQ pairs (\eg, ``\texttt{not green}'').

\noindent\textbf{CST.} Based on our general sample synthesizing mechanism CSS, we also propose a new CST training strategy for VQA, which helps the VQA models to focus on the critical objects and words, \ie, improving VQA models' visual-explainable and question-sensitive abilities. Specifically, the CST consists of two different training objectives: cross-entropy (XE) loss and contrastive loss (cf. Fig.~\ref{fig:2} (c)). For the XE training, we use CSS to generate a counterfactual sample for each original sample, and feed both two samples into the same VQA model. After training with these complementary samples simultaneously, VQA models are forced to focus on masked critical objects and words. For the contrastive training, we sample (or generate) a set of positive/negative samples for each original (anchor) sample, and propose two variants of \emph{supervised contrastive loss}~\cite{khosla2020supervised} for optimization. After training with the contrastive objective, the VQA models can further distinguish the original samples and superficially similar counterfactual ones.

For more effective contrastive training, we further design a novel positive and negative sample selection mechanism based on CSS. For each original training sample, we regard all samples with the same question type and ground-truth answers as \emph{positive} samples, and utilize CSS to generate corresponding counterfactual samples of each positive sample as \emph{negative} samples. Since CSS can effectively find the critical objects in the images or critical words in the questions, these positive and negative samples are superficially similar but semantically different, which is very effective for contrastive training.

\textbf{Contributions.} This paper is a substantial and systematic extension of our CVPR work~\cite{chen2020counterfactual}. Compared to the conference version, we have made several improvements: 
\begin{enumerate}[leftmargin=0.4cm]
    \item We propose an effective Counterfactual Samples Training (CST) strategy to train VQA models effectively. Specifically, we propose two variants of supervised contrastive loss for VQA, and designed a novel positive and negative sample selection mechanism based on CSS.
    
    \item We make two important modifications and improvements over the initial CSS~\cite{chen2019counterfactual}, which is denoted as CSS$^+$. Meanwhile, we have conducted a number of ablative studies to verify the effectiveness of these improvements.
    
    \item For more comprehensive evaluations, we evaluated CSST on three challenging datasets: VQA-CP v2, VQA-CP v1, and GQA-OOD. We achieve a new state-of-the-art performance on all benchmarks. Meanwhile, more experiments are conducted to further verify the generalization and superiority of CSST, including more datasets, backbones, ablation studies, and visualizations.
\end{enumerate}

Extensive ablations on VQA-CP and GQA-OOD benchmarks have demonstrated the effectiveness of CSST. Meanwhile, CSST can be seamlessly incorporated into any VQA architecture, which can not only improve their both visual-explainable and question-sensitive abilities, but also consistently boost their VQA performance. Particularly, by building on top of the model LMH+SAR~\cite{clark2019don,si2021check}, we achieve new record-breaking results on all three benchmarks.

\textbf{CSST vs. Existing Contrastive Traning Works in VQA}: To the best of our knowledge, there are a few works that utilize contrastive training in VQA~\cite{liang2020learning, kant2020contrast, zhu2020overcoming}. Especially for~\cite{liang2020learning}, which is also built on top of CSS. Specifically, they directly regard the counterfactual samples of an anchor sample in CSS as its negative samples, and compose critical objects/questions with original questions/images as corresponding positive samples. As shown in our experiments (cf. TABLE~\ref{tab:pos&neg}), this naive extension based on CSS extremely limits the diversity of training samples. In contrast, CSST can significantly improve the diversity of visual content and question in both positive and negative samples at contrastive training, which is empirically important (More details are discussed in Sec.~\ref{sec:3.3.2}). Meanwhile, we propose two variants of supervised contrastive loss for robust VQA.

\section{Related Work}

\noindent\textbf{Visual Question Answering (VQA).} VQA, \ie, understanding both the visual content and natural language question, and making the answer prediction, is an important research task in both computer vision and natural language processing. Benefits from the deep learning techniques and large-scale VQA datasets~\cite{antol2015vqa,goyal2017making,johnson2017clevr,hudson2019gqa}, VQA has realized impressive progress and achieved good results in real images. With the advance in large-scale multimodal representation pretraining, today's VQA performance is mainly dominated by pretrained multimodal BERT models~\cite{tan2019lxmert,su2020vl,li2020oscar,chen2020uniter}.

\noindent\textbf{Language Biases in VQA.}
Despite VQA is a typical multimodal task, a large body of research~\cite{jabri2016revisiting, agrawal2016analyzing, zhang2016yin, goyal2017making,niu2021introspective} has shown the existence of language biases in VQA. There are two main solutions to reduce language biases:

\textit{1. Balancing Datasets to Reduce Biases.} The most straightforward solution is to create more balanced datasets. For example, Zhang~\etal~\cite{zhang2016yin} and Goyal~\etal~\cite{goyal2017making} collected complementary images with opposite answers for all questions. Although these ``balanced'' datasets have reduced biases to some extent, the statistical biases from questions still can be leveraged~\cite{agrawal2018don}. As shown in VQA-CP, the performance of numerous models drop significantly compared to these ``balanced'' datasets. In this paper, CSST follows the same spirit of dataset balancing and trains VQA models with more complementary samples. Especially, we don't need any extra manual annotations.

\textit{2. Designing Models to Reduce Biases.} Another solution is to design specific debiasing models. So far, the most effective debiasing models for VQA are ensemble-based methods~\cite{ramakrishnan2018overcoming, grand2019adversarial, belinkov2019don, cadene2019rubi, clark2019don, mahabadi2019simple, han2021greedy}. In this paper, we propose a novel CSST strategy, which can be seamlessly incorporated into the ensemble-based models to further reduce the biases.

\noindent\textbf{Visual-Explainable Ability in VQA Models.}
To improve visual-explainable ability, early works~\cite{qiao2018exploring, liu2017attention, zhang2019interpretable} directly apply human attention as supervision to guide the models' attention maps. However, since the existence of strong biases, even with appropriate attention maps, the remaining layers of network may still disregard the visual signal~\cite{selvaraju2019taking}. Thus, some recent works~\cite{selvaraju2019taking, wu2019self} utilize Grad-CAM~\cite{selvaraju2017grad} to obtain the private contribution of each object to correct answers, and encourage the rank of all object contributions to be consistent with human annotations. Unfortunately, these models have two inherent drawbacks: 1) They need extra human annotations. 2) The training is not end-to-end.

\noindent\textbf{Question-Sensitive Ability in VQA Models.}
If VQA systems really ``understand'' the question, they should be sensitive to the linguistic variations in question. Surprisingly, to the best of our knowledge, there is only a few works~\cite{shah2019cycle} that have studied the influence of linguistic variations in VQA. Specifically, it designs a cycle-consistent loss between two dual tasks, and utilizes sampled noises to generate diverse questions. However, Shah~\etal~\cite{shah2019cycle} only consider the robustness of different rephrasings of the same question. In contrast, we also encourage models to perceive the difference in questions when changing some critical words.

\noindent\textbf{Data Augmentation in VQA.} Some concurrent works and following works after our CSS~\cite{chen2020counterfactual} also utilize data augmentation to improve VQA performance. Specifically, there are two directions: 1) \textit{Images Augmentation.} Almost all other image augmentation works~\cite{agarwal2019towards,pan2019question,gokhale2020mutant} resort to GAN~\cite{goodfellow2014generative} to generate corresponding counterfactual images, which is notorious for unstable training. Meanwhile, photo-realistic image generation itself is still an open challenge. 2) \textit{Questions Augmentation.} The most straightforward question augmentation strategy in VQA~\cite{kant2020contrast} is back-translation~\cite{edunov2018understanding}, which translating a sentence from one language to another and then translating it back using a pair of translation models. It is worth noting that some recent data augmentation works compose new training samples by directly pairing pristine images and questions from other samples~\cite{kil2021discovering,chen2022rethinking}.

In contrast, in this paper, our proposed CSST only masks critical objects or words, which is easier and more adoptable.

\noindent\textbf{Contrastive Training in VQA.} Contrastive learning techniques have achieved unprecedented success in vision community, especially for self-supervised representation learning~\cite{oord2018representation, he2020momentum, chen2020simple}. The core idea of contrastive learning is to maximize the mutual information between the input samples and positive samples, and minimize the one between negative samples. Currently, there are several VQA works~\cite{liang2020learning, kant2020contrast, zhu2020overcoming} that also utilize contrastive training to distinguish the original training samples and superficially similar counterparts. Different from all existing works, we equip our CSS mechanism with positive and negative selection for contrastive training, which not only increases the sample diversity, but also meets the semantic requirements.

\section{Approach}

We follow the common formulation and regard the VQA task as a multi-label classification problem. Without loss of generality, given a dataset $\mathcal{D} = \{I_i, Q_i, a_i \}^N_i$ consisting of triplets of images $I_i \in \mathcal{I}$, questions $Q_i \in \mathcal{Q}$ and answer sets $a_i \in \{a_i^j | a_i^j \in \mathcal{A}\}$, the VQA task learns a mapping $f_{vqa}: \mathcal{I} \times \mathcal{Q} \rightarrow [0, 1]^{|\mathcal{A}|}$, which produces an answer distribution given image-question pair. $\mathcal{I}$, $\mathcal{Q}$, and $\mathcal{A}$ denote the set of images questions, and answers, respectively. For each ground-truth answer $a_i^j \in a_i$, it has a soft target score $t_i^j$, and we use the same ground-truth soft target scores as in prior works~\cite{anderson2018bottom}. For simplicity, we omit subscript $i$ in the following sections.

In this section, we first introduce the UpDn baseline~\cite{anderson2018bottom} and a typical ensemble-based framework in Sec.~\ref{sec:3.1}. Then, we introduce the details of Counterfactual Samples Synthesizing (CSS) in Sec.~\ref{sec:3.2}. Last, we introduce the details of Counterfactual Samples Training (CST) in Sec.~\ref{sec:3.3}.

\subsection{Preliminaries} \label{sec:3.1}

\noindent\textbf{Bottom-Up Top-Down (UpDn) Model~\cite{anderson2018bottom}.}
For each image $I$, the UpDn uses an image encoder $e_v$ to output a set of object features: $\bm{V} = \{\bm{v}_1, ..., \bm{v}_{n_v}\}$, where $\bm{v}_i$ is $i$-th object feature. For each question $Q$, the UpDn uses a question encoder $e_q$ to output a set of word features: $\bm{Q} = \{\bm{w}_1, ..., \bm{w}_{n_q}\}$, where $\bm{w}_j$ is $j$-th word feature. Then, both $\bm{V}$ and $\bm{Q}$ are fed into the model $f_{vqa}$ to predict answer distributions:
\begin{equation} \label{eq:p_vqa}
P_{vqa}(\bm{a}|I, Q) = f_{vqa}(\bm{V}, \bm{Q}).
\end{equation}
Typically, model $f_{vqa}$ contains a soft attention mechanism, and it is trained with a (binary) cross-entropy (XE) loss.

\begin{algorithm}[t]
	\caption{Ensemble-based Model (fusion-based)}\label{alg:VQA}
    \textcolor{gray}{\textbf{Inputs:} original training sample ($I$, $Q$, $a$), $update\_cond$ is a control condition for parameter updates. \\
    \textbf{Outputs:} images features $\bm{V}$, question features $\bm{Q}$, and predicted answer distribution $P_{vqa}(\bm{a})$.}
	\begin{algorithmic}[1]
		\Function {\textbf{VQA}}{$I, Q, a, update\_cond$}
		\State $ \bm{V} \leftarrow e_v(I) $
		\State $ \bm{Q} \leftarrow e_q(Q) $
		\State $ P_{vqa}(\bm{a}) \leftarrow f_{vqa}(\bm{V}, \bm{Q}) $
		\If{$update\_cond$}
		\State $ P_{q}(\bm{a}) \leftarrow f_{q}(\bm{Q})$      \textcolor{gray}{\Comment{question-only model}}
		\State $ \hat{P}_{vqa}(\bm{a}) \leftarrow M\left(P_{vqa}(\bm{a}), P_{q}(\bm{a})\right)  $
		\State $ L_{\text{XE}} \leftarrow \text{XE}(\hat{P}_{vqa}(\bm{a}), a)$ 
		\textcolor{gray}{\Comment{update parameters}}
		\EndIf 
		\State \textbf{return} $\bm{V}, \bm{Q}, P_{vqa}(\bm{a})$
		\EndFunction
	\end{algorithmic}
\end{algorithm}

\noindent\textbf{Ensemble-Based Models.} As we discussed in Sec.~\ref{sec:intro}, the ensemble-based models can be grouped into two sub-types: \emph{adversary-based} and \emph{fusion-based}. Since adversary-based models~\cite{ramakrishnan2018overcoming, grand2019adversarial, belinkov2019don} always suffer severe unstable training and relatively worse performance, in this section, we only introduce the typical fusion-based framework~\cite{cadene2019rubi, clark2019don, mahabadi2019simple}. As shown in Algorithm~\ref{alg:VQA}, they introduce an auxiliary question-only model $f_q$ which takes $\bm{Q}$ as input and predicts answers:
\begin{equation}
P_{q}(\bm{a}|Q) = f_{q}(\bm{Q}).
\end{equation}

Then, they combine these two answer distributions and obtain a new answer distribution $\hat{P}_{vqa}(\bm{a})$ by a function $M$:
\begin{equation} \label{eq:eq3}
\hat{P}_{vqa}(\bm{a}|I, Q) = M \left( P_{vqa}(\bm{a}|I, Q), P_{q}(\bm{a}|Q) \right).
\end{equation}

In the training stage, the XE loss is computed based on the fused answer distribution $\hat{P}_{vqa}(\bm{a})$ and the training gradients are backpropagated through $f_{vqa}$ and $f_q$. In the test stage, only the model $f_{vqa}$ is used as plain VQA models.

\subsection{Counterfactual Samples Synthesizing (CSS)} \label{sec:3.2}

The overall structure of the CSS training scheme is shown in Algorithm~\ref{alg:css}. Specifically, for any \textbf{VQA} model, given an original training sample $(I, Q, a)$, CSS can generate two types of counterfactual samples: $(I^-, Q, a^-)$ by V-CSS or $(I, Q^-, a^-)$ by Q-CSS. In the following, we mainly introduce the details of V-CSS and Q-CSS. As shown in Algorithm~\ref{alg:css}, for each specific training sample, we only use one certain synthesizing mechanism, and $\delta$ is the trade-off weight (see Fig.~\ref{fig:5} (c) for more details about the influence of different $\delta$).

\subsubsection{V-CSS} \label{sec:v-css}

We sequentially introduce all steps of V-CSS following its execution path (\ie, line 5 to 8 in Algorithm~\ref{alg:css}), which consists of four main steps: initial objects selection (\textsc{IO\_Sel}), object local contributions calculation, critical objects selection (\textsc{CO\_Sel}), and dynamic soft answer assigning (\textsc{DSA\_Ass}).

\textbf{1. Initial Objects Selection (\textsc{IO\_Sel}).} In general, for any specific QA pair $(Q, a)$, only a few objects in image $I$ are related. To narrow the scope of critical object selection, we first construct a smaller object set $\mathcal{I}$, and assume all objects in $\mathcal{I}$ are possibly important in answering this specific question. Since we lack manual annotations about the critical objects for each sample, we followed~\cite{wu2019self} to extract objects which are highly related to the QA. Specifically, we first assign POS tags to each word in the QA using the spaCy POS tagger~\cite{honnibal2017spacy} and extract nouns in QA. Then, we calculate the cosine similarity between the GloVe~\cite{pennington2014glove} embeddings of object categories\footnote{For the VQA-CP dataset, we followed prior works~\cite{wu2019self} to utilize these object category annotations, and the influence of different sizes of initial objects are also illustrated in Fig.~\ref{fig:5} (a). For the GQA-OOD dataset, for fair comparisons with others, we skip this step for all experiments.} and the extracted nouns, the similarity scores between all objects in $I$ and the QA are denoted as $\mathcal{SIM}$. We select $|\mathcal{I}|$ objects with the highest $\mathcal{SIM}$ scores as the initial object set $\mathcal{I}$.

\begin{algorithm}[tbp]
	\caption{Counterfactual Samples Synthesizing (CSS)}\label{alg:css}
	\textcolor{gray}{\textbf{Inputs:} original training sample ($I$, $Q$, $a$), and a typical VQA model \textbf{VQA}.  \\
    \textbf{Outputs:} counterfactual training sample ($I^-$, $Q$, $a^-$) from V-CSS or  ($I$, $Q^-$, $a^-$) from Q-CSS.}
	\begin{algorithmic}[1]
		\Function {\textbf{CSS}}{$I, Q, a$}
		\State $ \bm{V}, \bm{Q}, P_{vqa}(\bm{a}) \leftarrow \textbf{\text{VQA}}(I, Q, a, \text{False})$
		\State $ cond \sim U[0, 1]$
		\If {$cond \geq \delta $}
		\textcolor{gray}{\Comment{execute V-CSS}}
		\State $ \mathcal{I} \leftarrow  \textsc{IO\_Sel}(I, Q) $
		\State $ s(a, \bm{v}_i) \leftarrow \mathcal{S}(P_{vqa}(a), \bm{v}_i)$
		\State $ I^+, I^- \leftarrow \textsc{CO\_Sel}(\mathcal{I}, \{s(a, \bm{v}_i) \}) $
		\State $ a^- \leftarrow \textsc{DSA\_Ass}(I^+, Q, \textbf{\text{VQA}}, a) $
		\State \textbf{return} ($I^-$, $Q$, $a^-$)
		\Else 
		\textcolor{gray}{\Comment{execute Q-CSS}}
		\State $ s(a, \bm{w}_i) \leftarrow \mathcal{S}(P_{vqa}(a), \bm{w}_i) $
		\State $ Q^+, Q^- \leftarrow \textsc{CW\_Sel}(\{s(a, \bm{w}_i)\})$
		\State $ a^- \leftarrow \textsc{DSA\_Ass}(I, Q^+, \textbf{\text{VQA}}, a) $
        \State \textbf{return} ($I$, $Q^-$, $a^-$)
		\EndIf
		\EndFunction
	\end{algorithmic}
\end{algorithm}

\textbf{2. Object Local Contributions Calculation.} After obtaining the initial object set $\mathcal{I}$, we start to calculate the local contribution of each object to the predicted probability of ground-truth answer. Following recent works~\cite{jain2019attention, selvaraju2019taking, wu2019self} which utilize modified Grad-CAM~\cite{selvaraju2017grad} to derive the local contribution of each participant, we calculate the contribution of $i$-th object feature to the ground-truth answer $a$ as:
\begin{equation} \label{eq:object_gradcam}
s(a, \bm{v}_i) = \mathcal{S}(P_{vqa}(a), \bm{v}_i) \coloneqq \left(\nabla_{\bm{v}_i} P_{vqa}(a) \right)^T\mathbf{1},
\end{equation}
where $P_{vqa}(a)$ is the predicted answer probability of the ground-truth answer a,  $\bm{v}_i$ is $i$-th object feature, and $\mathbf{1}$ is an all-ones vector. Obviously, if the score $s(a, \bm{v}_i)$ is higher, the contributions of object $\bm{v}_i$ to answer $a$ is larger.

\begin{figure}[tbp]
    \centering
    \includegraphics[width=0.95\linewidth]{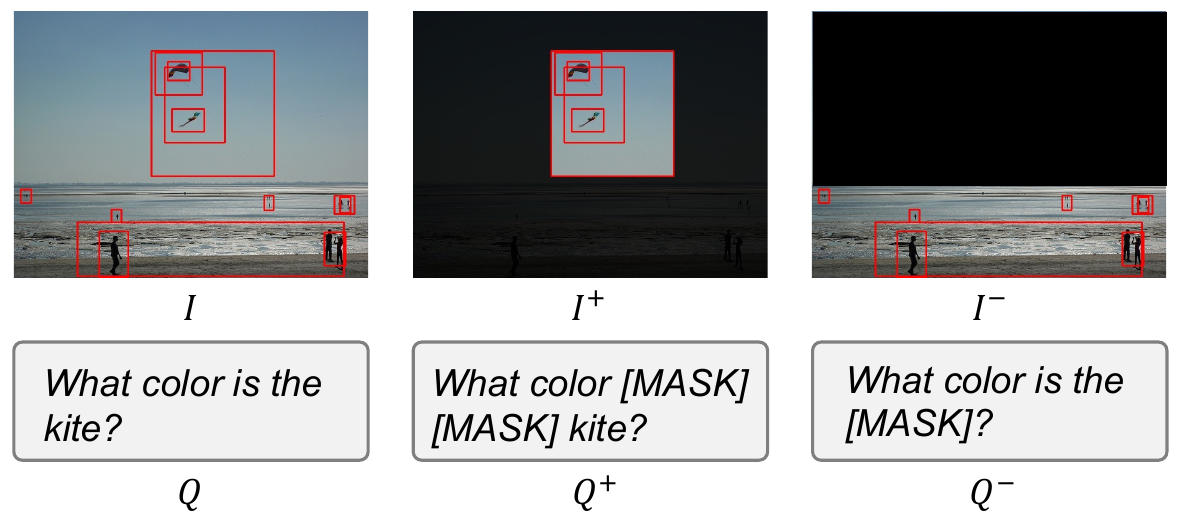}
    \vspace{-1em}
    \caption{An informal illustration example of the $I^+$, $I^-$, $Q^+$, and $Q^-$ in CSS. For $I^+$ and $I^-$, they are two mutual exclusive object sets. For $Q^+$ and $Q^-$, we show the example when word ``\texttt{kite}'' is selected as the critical word. For clarity, we omit some objects.}
    \label{fig:4}
\end{figure}

\textbf{3. Critical Objects Selection (\textsc{CO\_Sel}).} After obtaining the private contribution scores $s(a, \bm{v}_i)$ for all objects in $\mathcal{I}$, we select the top-K objects with highest scores as the critical object set $I^+$. The K is a dynamic number for each image, which is the smallest number meets Eq.~\eqref{eq:topk_objects}:
\begin{equation} \label{eq:topk_objects}
\sum_{\bm{v}_i \in I^+} \exp(s(a, \bm{v}_i)) / \sum_{\bm{v}_j \in \mathcal{I}} \exp(s(a, \bm{v}_j))  > \eta,
\end{equation}
where $\eta$ is a constant, we set $\eta=0.65$ in all experiments (See Fig.~\ref{fig:5} for more details about the influence of dynamic K setting). Since the visual features $\bm{V} = \{\bm{v}_i\}$ are extracted from a pretrained Faster R-CNN, \ie, they always have a lot of repetitions due to overlapped image regions. To avoid visual clue leakage from the other object visual features~\cite{lu202012}, we also regard the object that overlaps any critical objects by 60\% IoU or more as critical objects. Then, the counterfactual input $I^-$ is the absolute complement of set $I^+$ in set $I$, \ie, $I^- = I \backslash I^+$. We show an example of $I$, $I^+$ and $I^-$ in Fig.~\ref{fig:4}.

\begin{algorithm}[tbp]
	\caption{Dynamic Soft Answer Assigning (\textsc{DSA\_{Ass}})}\label{alg:daass}
	\textcolor{gray}{\textbf{Inputs:} VQ pair ($I^+$, $Q^+$), a typical VQA model \textbf{VQA}, and original ground-truth answer set $a$.  \\
    \textbf{Outputs:} automatically assigned pseudo gt answer set $a^-$, and corresponding ground-truth soft target scores $\{t^{i-}\}$.}
	\begin{algorithmic}[1]
		\Function {$\textsc{DA\_Ass}$}{$I^+, Q^+, \textbf{\text{VQA}}, a$}
		\State  $ \_, \_, P_{vqa}^+(\bm{a}) \leftarrow \textbf{\text{VQA}}(I^+, Q^+, a, \text{False}) $
		\State $ a^- \coloneqq a$
		\textcolor{gray}{\Comment{same gt answer set}}
		\For{$a^{i-} ~\text{in}~ a^-$}
		\State $t^{i-} \coloneqq 1 - P_{vqa}^+(a^{i-})$
		\textcolor{gray}{\Comment{soft answer}}
		\EndFor
		\State \textbf{return} $a^-$
		\EndFunction
	\end{algorithmic}
\end{algorithm}

\textbf{4. Dynamic Soft Answer Assigning (\textsc{DSA\_Ass}).} Given the counterfactual visual input $I^-$ and original question $Q$, we compose a new VQ pair ($I^-, Q$). To assign ground-truth answers for the VQ pair  ($I^-, Q$), we design a new dynamic soft answer assigning (\textsc{DSA\_Ass}) mechanism. The details of \textsc{DSA\_Ass} are shown in Algorithm~\ref{alg:daass}. Specifically, we first feed another VQ pair ($I^+, Q$) into the \textbf{VQA} model, and obtain the predicted answer distribution $P^+_{vqa}(\bm{a})$. We directly use the same set of ground-truth answer categories as the pseudo ground-truth for the corresponding counterfactual sample. Meanwhile, we re-assign the ground-truth probability for each answer category of $a^-$, \eg, the $i$-th answer category $a^{i-}$ is set to $P^{gt}(a^{i-}) = 1 - P_{vqa}^+(a^{i-})$. In an extreme case, if the model predicts all ground-truth answers correctly for the VQ pair ($I^+, Q$) with 100\% probabilities, then $a^-$ is a $\emptyset$, \ie, zero for all answer candidates. The basic motivation is that if the current model can predict ground-truth answers for ($I^+, Q$) (\ie, $I^+$ contains critical objects and $I^-$ not), the ground-truth for ($I^-, Q$) should not contain original ground-truth answers anymore, \eg, ``\texttt{not green}'' in Fig~\ref{fig:2}. Similarly, the higher confidence for ($I^+, Q$), the less probability for this answer to be a ground-truth for ($I^-, Q$).

\subsubsection{Q-CSS}
All steps in Q-CSS are similar to V-CSS. Following its execution path (\ie, line 11 to 13 in Algorithm~\ref{alg:css}), it consists of word local contribution calculation, critical words selection (\textsc{CW\_Sel}), and dynamic soft answer assigning (\textsc{DSA\_Ass}).

\textbf{1. Word Local Contribution Calculation.} Similar with the V-CSS (cf. Eq.~\eqref{eq:object_gradcam}), we calculate the contribution of $i$-th word feature to the ground-truth answer $a$ as:
\begin{equation} \label{eq:word_gradcam}
s(a, \bm{w}_i) = \mathcal{S}(P_{vqa}(a), \bm{w}_i) \coloneqq (\nabla_{\bm{w}_i} P_{vqa}(a))^T\mathbf{1}.
\end{equation}

\textbf{2. Critical Words Selection (\textsc{CW\_Sel}).}  In this step, we first extract question-type\footnote{We slightly abuse ``question-type'' here. For VQA-CP, the question-type denotes the default question-type annotations in the original dataset. For GQA-OOD, the question-type denotes the annotated local groups, which can be easily mapped back to each question. \label{qtype}} words for each question $Q$ (\eg, ``\texttt{what color}'' in Fig.~\ref{fig:4} is the question-type of question ``\texttt{What color is the kite?}''). Then, we select the top-K words with the highest scores from the remaining sentence (except question-type words) as the critical words. The counterfactual question $Q^-$ is the sentence by replacing all the critical words in $Q$ with a special token ``[\texttt{MASK}]''. Meanwhile, the $Q^+$ is the sentence by replacing all other words (except question-type and critical words) with ``[\texttt{MASK}]''. We show an example of $Q$, $Q^+$, and $Q^-$ in Fig.~\ref{fig:4}.

\textbf{3. Dynamic Soft Answer Assigning (\textsc{DSA\_Ass}.)} This step is identical to the \textsc{DSA\_Ass} in V-CSS, \ie, Algorithm~\ref{alg:daass}. For Q-CSS, the input for \textsc{DSA\_Ass} is the VQ pair $(I, Q^+)$.

\subsubsection{Highlights of Two Improvements on CSS~\cite{chen2020counterfactual}} \label{sec:3.2.3}
Compared to the initial CSS in the conference version~\cite{chen2020counterfactual}, we make two important modifications and improvements:
\begin{enumerate}
    \item \textbf{Object Overlapping in \textsc{CO\_Sel}.} Different from the word features in questions, which are relatively ``independent'', object features from the same image always have a lot of repetitions due to their overlapped image regions. We avoid this visual clue leakage by considering the IoU overlaps between all objects. This strategy helps models rely more on the calculated critical objects. 
    
    \item \textbf{``\emph{Soft}'' Answer Assigning in \textsc{DSA\_Ass}.} In initial CSS work~\cite{chen2020counterfactual}, we select top-K predictions based on $P^+_{vqa}$ (cf. Algorithm~\ref{alg:daass}), and assign ground-truth answers based on these top predictions. Compared to this ``hard'' answer selection, \textsc{DSA\_Ass} is more robust to the selection of hyperparameter $K$, and is more sensitive to perceive the changes of answer predictions (\ie, $P^+_{vqa}$).

\end{enumerate}

To distinguish these two types of CSS, we denote the current improved version as CSS$^+$ in the following sections (See results in Sec.~\ref{sec:exp} for comparisons between CSS \& CSS$^+$).

\subsection{Counterfactual Samples Training (CST)} \label{sec:3.3}

To further benefit from the counterfactual samples generated by CSS, we propose a CST strategy for model training (cf. Algorithm~\ref{alg:cst}). Specifically, it consists of a cross-entropy (XE) training loss and a contrastive (CR) training loss.
 
\subsubsection{XE Training on Counterfactual Samples}

For XE training, we directly follow other state-of-the-art VQA models (\eg, UpDn~\cite{anderson2018bottom}) and use binary cross-entropy loss as training objective, \ie, 
\begin{equation}
    \begin{split}
        L_{\text{XE}} & = - \sum^M_i \left[t^i \log \sigma(\hat{P}_{vqa}(\bm{a}|I, Q)) + \right. \\
        & \left. + (1 - t^i) \log (1 - \sigma(\hat{P}_{vqa}(\bm{a}|I, Q))) \right],
    \end{split}
\end{equation}
where $\sigma$ denotes the sigmoid activation function, and $t^i$ is the soft target score of $i$-th ground-truth answer for this training sample (cf. line 5 in Algorithm~\ref{alg:daass}). Different from the most prevalent VQA framework which only uses original samples in the XE training (cf. Fig.~\ref{fig:2} (a)), in our XE training, we feed both original training samples and their corresponding counterfactual samples into the same \textbf{VQA} model (\ie, line 3 to 5 in Algorithm~\ref{alg:cst}).

\begin{algorithm}[tbp]
    \textcolor{gray}{\textbf{Inputs:} original training sample ($I$, $Q$, $a$), and a typical VQA model \textbf{VQA}, and \textbf{CSS}.  
    }
	\caption{Counterfactual Samples Training (CST)}\label{alg:cst}
	\begin{algorithmic}[1]
		\Function {\textbf{CST}}{$I, Q, a$} 
        \State \textcolor{gray}{\# XE training} 
        \State $(I^-, Q, a^-) \leftarrow \textbf{\text{CSS}}(I, Q, a)$
		\textcolor{gray}{\Comment{V-CSS for example}}
        \State $ \_, \_, P_a \leftarrow \textbf{\text{VQA}}(I, Q, a, \text{True})$
        \textcolor{gray}{\Comment{original samples}}
        \State $ \_, \_, \_ \leftarrow \textbf{\text{VQA}}(I^-, Q, a^-, \text{True})$
        \textcolor{gray}{\Comment{counter. samples}}
        \\
        \State \textcolor{gray}{\# samples selection} 
		\State $(I_p, Q_p, a) \leftarrow \textsc{POS\_Sel}(I, Q, a) $
		\textcolor{gray}{\Comment{pos. samples}}
		\State $\{(I_n^i, Q_n^i, \_)\} \leftarrow \textsc{NEG\_Sel}(I, Q, a) $ 
        \textcolor{gray}{\Comment{neg. samples}}
        \\
        \State \textcolor{gray}{\# contrastive training}
        \State $\_, \_, P_p \leftarrow \textbf{\text{VQA}} (I_p, Q_p, _, \text{False})$
        \textcolor{gray}{\Comment{for pos. samples}}
        \State $\_, \_, P^i_n \leftarrow \textbf{\text{VQA}} (I_n^i, Q_n^i, _, \text{False})$
        \textcolor{gray}{\Comment{for neg. samples}}
        \State $L_{\text{CR}} \leftarrow \text{CR}(P_a, P_p, \{P_n^i\})$
        \textcolor{gray}{\Comment{CR training}}
		\EndFunction
	\end{algorithmic}
\end{algorithm}

\subsubsection{Contrastive Training on Counterfactual Samples} \label{sec:3.3.2}
For contrastive training, we propose a novel and effective positive and negative samples selection mechanism based on our CSS, and two variants of contrastive loss for VQA.

\noindent\textbf{Positive \& Negative Samples Selection Strategies.} Without loss of generality, we regard each original training sample as the \emph{anchor} sample, and we regard all samples with the same question-type\footref{qtype} and ground-truth answer categories as the candidate positive set\footnote{For those samples do not have any sample meets the two requirements, we directly regard the original sample as the candidate positive set, and the proportions are quite small, \eg, $\approx$ 1\% for VQA-CP v2.}. In each training step, we randomly sample one sample from the candidate set as the \emph{positive} sample (\ie, \textsc{POS\_Sel} in Algorithm~\ref{alg:cst}). For each positive sample, we compose four different types of \emph{negative} samples, which can be categorized into two groups based on their sampling strategies. The first group of negative samples are the counterfactual samples of the positive samples corresponding to V-CSS and Q-CSS, respectively. The second group of negative samples are randomly sampled samples: including 1) a random sampled sample with the same question-type\footref{qtype} but different ground-truth answer; and 2) a composed sample by replacing the original image with a randomly sampled image from the same batch. All four types of negative sample selection strategies constitute the \textsc{NEG\_Sel} step in Algorithm~\ref{alg:cst}. Detailed quantitative results of different samples are reported in TABLE~\ref{tab:pos&neg}.

\emph{Advantages over Existing Positive/Negative Sampling Solutions.} Currently, there are two types of positive and negative sample selection strategies. The first type also builds on top of our CSS~\cite{liang2020learning}. For each sample $(I, Q)$, they directly use the $(I^-, Q)$ and $(I, Q^-)$ in our \textbf{CSS} (cf. Algorithm~\ref{alg:css}) as negative samples. The second type is directly randomly replaced an image or question~\cite{kant2020contrast} to compose negative samples (\eg, $(I_{\text{rand}}, Q)$ or $(I, Q_{\text{rand}})$). Instead, we benefit from the ability of CSS, and generate counterfactual samples for positive samples (our first group), \ie, negative samples consist of $(I_{pos}^-, Q_{pos})$ and $(I_{pos}, Q_{pos}^-)$. Compared to the first solution, our strategy significantly increases the \emph{diversity} of visual contents and questions in positive and negative samples, \ie, in each training epoch, we can use totally different counterfactual samples as negative samples. Compared to the second solution, our strategy based on CSS can help models to focus more on fine-grained differences (masked critical objects and words) under contrastive training.

\noindent\textbf{Two Variants of Contrastive Loss for VQA.} In this paper, we propose two different types of contrastive loss for VQA: a global version which calculates the similarity (or distance) between anchor samples and positive/negative samples on the whole answer probabilities (dubbed as CR-G), and a local version which calculates the distance based on only ground-truth answer probabilities (dubbed as CR-L).

\emph{Global Contrastive Loss (CR-G).} Given the predicted answer distributions (before the sigmoid activation function) of anchor sample, positive sample, and negative samples (denoted as $P_a$, $P_p$ and $\{P_n^i\}$ respectively), the CR-G loss is:
\begin{equation}
    L_{\text{CR-G}} = - \log \left( 
    \frac{\mathrm{e}^{s(P_a, P_p)/\tau}}{\mathrm{e}^{s(P_a, P_p)/\tau} + \sum_i\mathrm{e}^{s(P_a, P_n^i)/\tau}} 
    \right),
\end{equation}
where $s(P_i, P_j)$ is the cosine similarity of answer distributions $P_i$ and $P_j$, and $\tau$ is the hyperparameter temperature.

\emph{Local Contrastive Loss (CR-L).} Instead of calculating similarity based on the whole answer distributions, there is another alternative choice that only focuses on probabilities of ground-truth answers. Thus, the CR-L loss is:
\begin{equation} \label{eq:9}
    L_{\text{CR-L}} = - \log  \left( 
    \frac{\mathrm{e}^{\tilde{P}_a(m)/\tau}}{\mathrm{e}^{\tilde{P}_a(m)/\tau} + \sum_i \tilde{P}_n^i(m) \times \mathrm{e}^{\tilde{P}_n^i(m)/\tau} } 
\right),
\end{equation}
where $m$ is the answer index of ground-truth answer with the highest target score, and $\tilde{P}_*(\cdot)$ is the answer probabilities after sigmoid activation function (\ie, $\tilde{P}_*(\cdot) = \sigma(P_*(\cdot))$). In addition, we add $\tilde{P}_n^i(m)$ as weights which helps models to focus more on hard negative samples with larger $\tilde{P}_n^i(m)$.

\emph{Differences between CR-G and CR-L.} For CR-L, since it only optimizes the VQA models along the single ground-truth answer direction (\ie, the $m$ answer index in Eq.~\eqref{eq:9}, it only suppresses the predicted scores of negative samples on these original ground-truth answers. However, for vanilla VQA models without any debiasing techniques (\eg, UpDn), their predicted scores of these negative samples (biased samples) on the original ground-truth answers are over-high. Thus, the CR-L training objectives can significantly improve the performance of these vanilla VQA models. In contrast, for debiasing VQA models (\eg, ensemble-based model LMH), they always have somewhat debiasing ability, and their predicted scores for these original ground-truth answers are not so high. Thus, for these debiasing models, it would be better to use CR-G, which directly calculates the sample similarity based on the global answer distributions.

\begin{figure*}[htbp]
    \centering
    \includegraphics[width=0.95\linewidth]{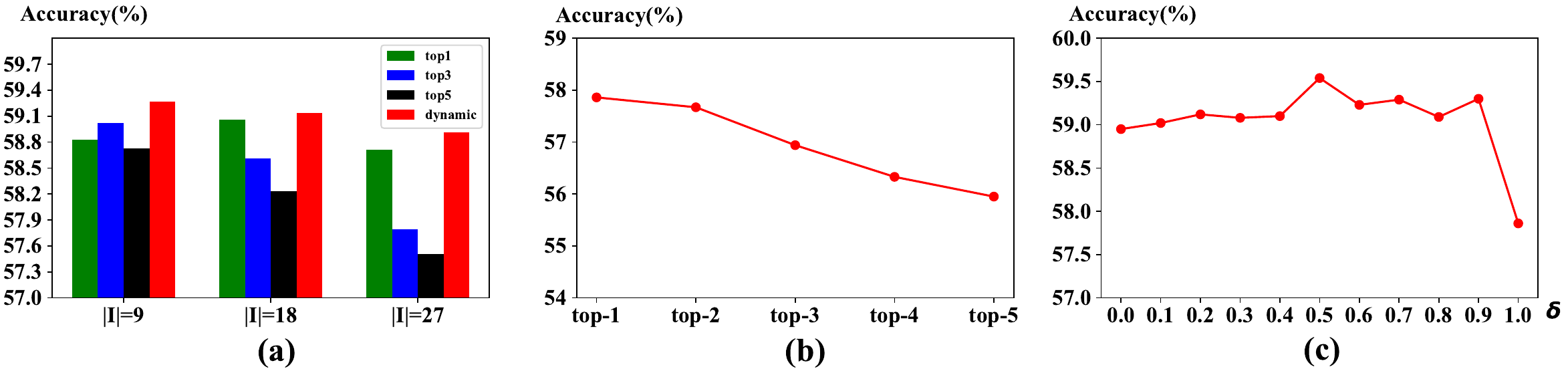}
    \vspace{-1em}
    \caption{\textbf{Ablations}. Accuracies (\%) on VQA-CP v2 test set of different hyperparameters settings of V-CSS$^+$ or Q-CSS$^+$. (a) The results of different sizes of $\mathcal{I}$ and critical objects in V-CSS$^+$. All results come from model LMH+V-CSS$^+$. (b) The results of different sizes of critical words in Q-CSS$^+$. All results come from model LMH+Q-CSS$^+$. (c) The results of different proportions $\delta$ between V-CSS$^+$ and Q-CSS$^+$. All results come from model LMH+V-CSS$^+$+Q-CSS$^+$. LMH denotes a prevalent debiasing VQA model~\cite{clark2019don}.}
    \label{fig:5}
\end{figure*}

\section{Experiments} \label{sec:exp}

\subsection{Datasets and Evaluation Metrics}
\noindent\textbf{Datasets.} We evaluated CSST on three challenging diagnostic VQA benchmarks: 1) \textbf{VQA-CP}~\cite{agrawal2018don}. It is a re-organization of the training and test sets of widely-used VQA v1~\cite{antol2015vqa} and VQA v2~\cite{goyal2017making}, where the answer distribution of each question type in the training set is made explicitly different from the one in the test set. There are two VQA-CP datasets: VQA-CP v2 and VQA-CP v1. We followed the official splits~\cite{agrawal2018don} for both two datasets. We also reported our results on the VQA v2 validation set for more complete comparisons. 2) \textbf{GQA-OOD}~\cite{kervadec2021roses}. It is a fine-grained re-organization of GQA~\cite{hudson2019gqa}. They share the same training set, but GQA-OOD introduces fine-grained shifts into both the validation and test sets.

\noindent\textbf{Evaluation Metrics.}
For evaluating the model accuracy on VQA-CP v2/v1, we followed standard metric~\cite{antol2015vqa}. Similarly, we reported accuracy on all test samples (\textbf{All}) and three different categories separately: Yes/No (\textbf{Y/N}), number counting (\textbf{Num}), and other (\textbf{Other}) categories. Particularly, the ``Y/N'', ``Num'', and ``Other'' types denote the set of questions that are answered ``yes/no'', ``number'', and all other types of answers (neither answered ``yes/no'' nor ``number''), respectively. For evaluating the model accuracy on GQA-OOD, there are four metrics: 1) \textbf{Acc-A} (all): overall accuracy over all test samples. 2) \textbf{Acc-T} (tail): the accuracy on OOD samples (\ie, samples of the tail of the answer class distribution). 3) \textbf{Acc-H} (head): the accuracy on in-domain samples (\ie, samples of the head of each local group). 4) $\mathbf{\Delta}$ = (Acc-H - Acc-T) / Acc-T: the error prediction imbalance between frequent (in-domain) and rare (OOD) answers.

\subsection{Experimental Settings} \label{sec:expsetting}
\noindent\textbf{Data Preprocessing Details.}
For fair comparisons, we did all the same data preprocessing steps with the widely-used UpDn model~\cite{anderson2018bottom} using the publicly available reimplementation\footnote{\href{https://github.com/hengyuan-hu/bottom-up-attention-vqa}{https://github.com/hengyuan-hu/bottom-up-attention-vqa}\label{updn_github}}. Specifically, for image feature extraction, we used Faster R-CNN~\cite{ren2015faster} pretrained on Visual Genome to extract the top-$K$ objects. Following the convention of prior works, we set $K=36$ for VQA-CP (and VQA v2), and $K=100$ for GQA-OOD. For question feature extraction, there are slight differences for the models with different backbones. For UpDn-based models, we set the maximum length of questions as 14, and the word embeddings are initialized with GloVe ~\cite{pennington2014glove} vectors with a dimension of 300. A single-layer GRU is used to obtain question embedding vectors with a dimension of 1,024. For LXMERT-based models, we used the official tokenizer in LXMERT to segment each question into word tokens. A pre-trained LXMERT model is used to obtain question features with a dimension of 768.

\noindent\textbf{Training Details and Hyperparameters.}
We trained UpDn-based models or LXMERT-based models for 30 epochs with batch size 512 on all datasets. We used the Adamax algorithm as the optimizer following the public reimplementation\footref{updn_github}. All parameters were initialized from scratch and the random seed was set to 0. Loss weights of $L_\text{XE}$, $L_\text{CR-G}$, and $L_\text{CR-L}$ were set to 1, 1, and 8 in most of experiments. $\tau$ in the contrastive loss was set to 1 in all experiments. Results about the influence of hyperparameters are reported in Sec.~\ref{sec:4.3.1}


\begin{table*}[tbp]
    \caption{Accuracies (\%) of different VQA architectures on VQA-CP v2 test set and GQA-OOD testdev set. Results are shown for ``all'' (All) questions, ``yes/no'' (Y/N) questions, ``number'' (Num) questions, and ``other'' (Other) questions. CSST-G and CSST-L denote the CSST with CR-G and CR-L losses, respectively. $^*$ denotes the results from our reimplementation using official codes. Q-CSS$^+$/V-CSS$^+$/CSS$^+$ denote the results of this improved version of CSS, and Q-CSS/V-CSS/CSS denote the results from~\cite{chen2020counterfactual}.}
	\vspace{-0.5em}
	\subfloat[Ablation studies of different CSST variants and components on UpDn~\cite{anderson2018bottom} base model.]{
		\tablestyle{3.5pt}{1.05}\begin{tabular}{|l|x{23}x{23}x{23}x{23}|}
			\hline
			\multirow{2}{*}{Models} & \multicolumn{4}{c|}{VQA-CP v2} \\
			& All & Y/N & Num & Other  \\
			\hline
			\textbf{UpDn}~\cite{anderson2018bottom}  & 39.74 & 42.27 & 11.93  & 46.05 \\
			Baseline$^*$ & 39.68 & 41.93 & 12.68 & 45.91 \\
			~+Q-CSS & 40.05 & 42.16 & 12.30 & 46.56 \\
			~+V-CSS & 40.98 & 43.12 & 12.28 & 46.86 \\
			~+CSS & 41.16 & 43.96 & 12.78 & 47.48 \\
			~+CSS$^+$ & \cellcolor{mygray-bg}{40.84} & \cellcolor{mygray-bg}{43.09}  & \cellcolor{mygray-bg}{12.74}  & \cellcolor{mygray-bg}{47.37} \\
			~+CSST-G & \cellcolor{mygray-bg}{41.68} & \cellcolor{mygray-bg}{45.70}  & \cellcolor{mygray-bg}{14.01}  & \cellcolor{mygray-bg}{47.16} \\
			~+CSST-L & \cellcolor{mygray-bg}{\textbf{56.55}} & \cellcolor{mygray-bg}{\textbf{80.45}}  & \cellcolor{mygray-bg}{\textbf{36.29}}  & \cellcolor{mygray-bg}{\textbf{49.58}}  \\
		    \hline
			\multirow{2}{*}{Models} & \multicolumn{4}{c|}{GQA-OOD} \\
			& Acc-A & Acc-T & Acc-H & $\Delta$$\downarrow$  \\
			\hline
			\textbf{UpDn}~\cite{anderson2018bottom}  & 46.4 & 42.1 & 49.1 & 16.6 \\
			Baseline$^*$ & 46.96 & 42.90 & 49.45 & 15.3 \\
			~+Q-CSS$^+$ & \cellcolor{mygray-bg}{46.42} & \cellcolor{mygray-bg}{43.37}  & \cellcolor{mygray-bg}{48.30}  & \cellcolor{mygray-bg}{11.4} \\
			~+V-CSS$^+$ & \cellcolor{mygray-bg}{46.64} & \cellcolor{mygray-bg}{43.56}  & \cellcolor{mygray-bg}{48.53}  & \cellcolor{mygray-bg}{11.4} \\
            ~+CSS & \cellcolor{mygray-bg}{47.03} & \cellcolor{mygray-bg}{43.27}  & \cellcolor{mygray-bg}{49.34}  & \cellcolor{mygray-bg}{14.0} \\
			~+CSS$^+$ & \cellcolor{mygray-bg}{46.17} & \cellcolor{mygray-bg}{\textbf{44.40}}  & \cellcolor{mygray-bg}{47.26}  & \cellcolor{mygray-bg}{\textbf{6.4}} \\
			~+CSST-G & \cellcolor{mygray-bg}{\textbf{47.78}} & \cellcolor{mygray-bg}{44.21}  & \cellcolor{mygray-bg}{\textbf{49.97}}  & \cellcolor{mygray-bg}{13.0} \\
			~+CSST-L & \cellcolor{mygray-bg}{47.25} & \cellcolor{mygray-bg}{43.37}  & \cellcolor{mygray-bg}{49.62}  & \cellcolor{mygray-bg}{14.4} \\
			\hline
	\end{tabular}}\hspace{2mm}
	\subfloat[Ablation studies of different CSST variants and components on RUBi~\cite{cadene2019rubi} base model.]{
		\tablestyle{3.5pt}{1.05}\begin{tabular}{|l|x{23}x{23}x{23}x{23}|}
			\hline
			\multirow{2}{*}{Models} & \multicolumn{4}{c|}{VQA-CP v2} \\
			& All & Y/N & Num & Other  \\
			\hline
			\textbf{RUBi}~\cite{cadene2019rubi} & 44.23 & --- & ---  & ---  \\
			Baseline$^*$ & 45.23 & 64.85 & 11.83 & 44.11 \\
			~+Q-CSS & 46.31 & 68.70 & 12.15 & 43.95 \\
			~+V-CSS & 46.00 & 62.08 & 11.84 & \textbf{46.95} \\
			~+CSS & 46.67 & 67.26 & 11.62 & 45.13 \\
			~+CSS$^+$ & \cellcolor{mygray-bg}{47.46} & \cellcolor{mygray-bg}{\textbf{69.42}}  & \cellcolor{mygray-bg}{12.18}  & \cellcolor{mygray-bg}{45.63} \\
			~+CSST-G & \cellcolor{mygray-bg}{47.06} & \cellcolor{mygray-bg}{65.91}  & \cellcolor{mygray-bg}{12.61}  & \cellcolor{mygray-bg}{46.64} \\
			~+CSST-L & \cellcolor{mygray-bg}{\textbf{47.65}} & \cellcolor{mygray-bg}{62.06}  & \cellcolor{mygray-bg}{\textbf{33.46}} & \cellcolor{mygray-bg}{43.98} \\
		    \hline
			\multirow{2}{*}{Models} & \multicolumn{4}{c|}{GQA-OOD} \\
			& Acc-A & Acc-T & Acc-H & $\Delta$$\downarrow$  \\
			\hline
			\textbf{RUBi}~\cite{cadene2019rubi}  & 38.8 & 35.7 & 40.8 & 14.3 \\
			Baseline$^*$ & 45.85 & 43.37 & 47.37 & \textbf{9.2} \\
			~+Q-CSS$^+$ & \cellcolor{mygray-bg}{47.32} & \cellcolor{mygray-bg}{43.18} & \cellcolor{mygray-bg}{49.86} & \cellcolor{mygray-bg}{15.5} \\
		    ~+V-CSS$^+$ & \cellcolor{mygray-bg}{45.46} & \cellcolor{mygray-bg}{41.86} & \cellcolor{mygray-bg}{47.66} & \cellcolor{mygray-bg}{13.9} \\
            ~+CSS & \cellcolor{mygray-bg}{46.39} & \cellcolor{mygray-bg}{42.05}  & \cellcolor{mygray-bg}{49.05}  & \cellcolor{mygray-bg}{16.6} \\
			~+CSS$^+$ & \cellcolor{mygray-bg}{46.75} & \cellcolor{mygray-bg}{42.90}  & \cellcolor{mygray-bg}{49.11}  & \cellcolor{mygray-bg}{14.5} \\
			~+CSST-G & \cellcolor{mygray-bg}{47.64} & \cellcolor{mygray-bg}{43.18}  & \cellcolor{mygray-bg}{50.38}  & \cellcolor{mygray-bg}{16.7} \\
			~+CSST-L & \cellcolor{mygray-bg}{\textbf{48.39}} & \cellcolor{mygray-bg}{\textbf{44.31}}  & \cellcolor{mygray-bg}{\textbf{50.89}}  & \cellcolor{mygray-bg}{14.8} \\
			\hline
	\end{tabular}}\hspace{2mm}
	\subfloat[Ablation studies of different CSST variants and components on LMH~\cite{clark2019don} base model.]{
		\tablestyle{3.5pt}{1.05}\begin{tabular}{|l|x{23}x{23}x{23}x{23}|}
			\hline
			\multirow{2}{*}{Models} & \multicolumn{4}{c|}{VQA-CP v2} \\
			& All & Y/N & Num & Other  \\
			\hline
			\textbf{LMH}~\cite{clark2019don}  & 52.05 & --- & ---  & ---  \\
			Baseline$^*$ & 52.45 & 69.81 & 44.46 & 45.54 \\
			~+Q-CSS & 56.66 & 80.82 & 45.83 & 46.98 \\
			~+V-CSS & 58.23 & 80.53 & 52.48 & 48.13 \\
			~+CSS & 58.95 & 84.37 & 49.42 & 48.21 \\
			~+CSS$^+$ & \cellcolor{mygray-bg}{59.54} & \cellcolor{mygray-bg}{83.37}  & \cellcolor{mygray-bg}{52.57}  & \cellcolor{mygray-bg}{48.97} \\
			~+CSST-G & \cellcolor{mygray-bg}{\textbf{61.66}} & \cellcolor{mygray-bg}{\textbf{90.20}}  & \cellcolor{mygray-bg}{\textbf{54.42}}  & \cellcolor{mygray-bg}{48.69} \\
			~+CSST-L & \cellcolor{mygray-bg}{55.38} & \cellcolor{mygray-bg}{68.01}  & \cellcolor{mygray-bg}{52.89}  & \cellcolor{mygray-bg}{\textbf{49.45}} \\
		    \hline
			\multirow{2}{*}{Models} & \multicolumn{4}{c|}{GQA-OOD} \\
			& Acc-A & Acc-T & Acc-H & $\Delta$$\downarrow$  \\
			\hline
			\textbf{LMH}~\cite{clark2019don} & --- & --- & --- & --- \\
			Baseline$^*$ & 43.96 & 40.73 & 45.93 & 12.8 \\
			~+Q-CSS$^+$ & \cellcolor{mygray-bg}{43.45} & \cellcolor{mygray-bg}{41.20} & \cellcolor{mygray-bg}{44.84} & \cellcolor{mygray-bg}{8.8} \\
			~+V-CSS$^+$ & \cellcolor{mygray-bg}{42.63} & \cellcolor{mygray-bg}{41.02} & \cellcolor{mygray-bg}{43.62} & \cellcolor{mygray-bg}{6.3} \\
            ~+CSS & \cellcolor{mygray-bg}{43.10} & \cellcolor{mygray-bg}{41.49}  & \cellcolor{mygray-bg}{44.09}  & \cellcolor{mygray-bg}{\textbf{6.3}} \\
			~+CSS$^+$ & \cellcolor{mygray-bg}{44.24} & \cellcolor{mygray-bg}{41.20}  & \cellcolor{mygray-bg}{46.11}  & \cellcolor{mygray-bg}{11.9} \\
			~+CSST-G & \cellcolor{mygray-bg}{45.42} & \cellcolor{mygray-bg}{42.90}  & \cellcolor{mygray-bg}{46.97}  & \cellcolor{mygray-bg}{9.5} \\
			~+CSST-L & \cellcolor{mygray-bg}{\textbf{47.93}} & \cellcolor{mygray-bg}{\textbf{44.31}}  & \cellcolor{mygray-bg}{\textbf{50.14}}  & \cellcolor{mygray-bg}{13.2} \\
			\hline
	\end{tabular}}\hspace{3mm}
	\vspace{-1em}
	\label{tab:boost_models}
\end{table*}

\addtolength{\tabcolsep}{-1pt}
\begin{table}[tbp]
	\small
	\caption{Ablation (\%) studies of the influence of different positive and negative samples selection strategies on VQA-CP v2. Results are shown for ``all'' (All) questions, ``yes/no'' (Y/N) questions, ``number'' (Num) questions, and ``other'' (Other) questions. ``CSS'' denotes the model using only two negative samples from V-CSS$^+$ and Q-CSS$^+$ in contrastive (CR) training. ``Rand'' denotes the model using only the last two randomly composed negative samples in the CR training. $^*$ denotes the results from our reimplementation.}
	\vspace{-1em}
	\begin{center}
		\scalebox{0.98}{
			\begin{tabular}{| l | c c | c c c c|}
				\hline
				Models & CSS & Rand & All & Y/N & Num & Other  \\
				\hline
				Baseline (LMH) & & & 52.45 & 69.81 & 44.46 & 45.54 \\
				LMH-CSS & & & 58.95 & 84.37 & 49.42 & 48.21 \\
				~+CL~\cite{liang2020learning} & & & 59.18 & 86.99 & 49.89 & 47.16 \\
				LMH-CSS$^+$ & & & 59.54 & 83.37 & 52.57 & 48.97  \\
				~+CL$^*$~\cite{liang2020learning} & & & 59.78 & 83.85 & 53.12 & \textbf{49.00} \\
				~+CR & & \boldcheckmark & \cellcolor{mygray-bg}{60.74} & \cellcolor{mygray-bg}{89.49} & \cellcolor{mygray-bg}{52.05} & \cellcolor{mygray-bg}{48.06} \\
				~+CR & \boldcheckmark & & \cellcolor{mygray-bg}{61.13} & \cellcolor{mygray-bg}{90.09} & \cellcolor{mygray-bg}{53.48} &  \cellcolor{mygray-bg}{48.06} \\
				~+CR (CSST-G) & \boldcheckmark & \boldcheckmark & \cellcolor{mygray-bg}{\textbf{61.66}} & \cellcolor{mygray-bg}{\textbf{90.20}} & \cellcolor{mygray-bg}{\textbf{54.42}} & \cellcolor{mygray-bg}{48.69} \\
				\hline
				\textcolor{gray}{~+CL$^*$+CR}& \textcolor{gray}{\boldcheckmark} & \textcolor{gray}{\boldcheckmark} & \textcolor{gray}{61.36} & \textcolor{gray}{89.03} & \textcolor{gray}{53.65} & \textcolor{gray}{48.98} \\
				\hline
			\end{tabular}
		} 
	\end{center}
	\label{tab:pos&neg}
\end{table}
\addtolength{\tabcolsep}{1pt}

\noindent\textbf{Adapting Ensemble-based Methods to GQA-OOD Benchmark.} In original GQA-OOD paper~\cite{kervadec2021roses}, the authors claim none of the typical ensemble-based models (\eg, RUBi~\cite{cadene2019rubi} and LMH~\cite{clark2019don}) can improve the Acc-T performance on GQA-OOD, and they even deteriorate Acc-H performance. In this paper, we argue that it is not suitable to apply these debiasing methods on all samples, because ``biases'' only reside in \emph{imbalanced local groups}~\cite{kervadec2021roses}. Therefore, in our experiments, we only utilized these debiasing methods on imbalanced local groups. Since the local groups of samples are different in the training and test sets, in the training stage, we only selected the imbalanced local groups from the training set based on the same Shannon entropy and threshold, \ie, we don't use any extra information about the test set in advance.

\noindent\textbf{Adapting LXMERT Backbone for VQA.} LXMERT~\cite{tan2019lxmert} itself is a multimodal BERT model, which can be easily used for VQA. Similar to other multimodal BERT models, after pretraining, it typically utilizes the output of the [CLS] token as a multimodal fused feature, and trains a linear classifier for answer prediction. Unfortunately, this setting is slightly different from the UpDn backbone, whose inputs are both visual and question features. To benefit from the pretrained weights and seamlessly equipped LXMERT into other existing ensemble-based methods with UpDn backbone, in this paper, we utilized the outputs of visual and language embedding tokens as the visual and question features, respectively (\ie, replacing original inputs of ensemble-based methods). Pretrained weights of LXMERT are kept fixed. (See results of models with LXMERT in following sections.)

\noindent\textbf{Adapting SAR~\cite{si2021check} strategy to LMH-CSST.} In our experiments, we also equipped the model-agnostic SAR~\cite{si2021check} to our LMH-CSST to further improve performance. Specifically, for the Select And Rerank (SAR) module, we used SAR to refer to SAR+LMH (\ie, incorporate LMH into SAR). We chose LMH-CSST as Candidate Answer Selector (CAS), and used top-20 answers as candidates. We utilized the ``R $\to$ C'' strategy to combine question and answer into a synthetic dense caption. We trained SAR for 10 epochs, and batch size was set to 32. For Question Type Discriminator, we selected 1 or 2 candidates for Y/N questions and 12 candidate answers for non-Y/N questions when testing on VQA-CP. Meanwhile, Since we lack manual question type annotations for samples on GQA-OOD, we removed Question Type Discriminator when testing on GQA-OOD, and selected 12 candidates for all questions. We refer readers to SAR ~\cite{si2021check} for more details.


\begin{table*}
	\small
	\caption[]{Accuracies (\%) on VQA-CP v2 test set and VQA v2 val set of state-of-the-art models. Results are shown for ``all'' (All) questions, ``yes/no'' (Y/N) questions, ``number'' (Num) questions, and ``other'' (Other) questions. The gap (Other) represents the accuracy difference between the Other category of VQA v2 and VQA-CP v2. ``Pre.'' represents ``Pretraining'', \ie, these models use extra datasets at their pretraining stage (\eg, LXMERT backbone~\cite{tan2019lxmert} or SAR~\cite{si2021check}). ``Ann.'' represents ``Annotation'', \ie, these models rely on extra manual annotations. $^\dagger$ represents the \emph{ensemble-based} methods, $^*$ indicates the results from our reimplementation using official codes, $^\ddag$ denotes these models use the same adapting strategy for LXMERT backbone as in Sec.~\ref{sec:expsetting}.} 
	\vspace{-1em}
	\begin{center}
		\scalebox{0.96}{
			\begin{tabular}{| l | c | c | c | c c c c | c c c c| c |}
				\hline
				\multirow{2}{*}{Models}  & \multirow{2}{*}{Base} & \multirow{2}{*}{Pre.} & \multirow{2}{*}{Ann.} & \multicolumn{4}{c|}{VQA-CP v2 test $\uparrow$} & \multicolumn{4}{c|}{VQA v2 val $\uparrow$} & Gap $\downarrow$ \\
				& & & & All & Yes/No & Num & Other & All & Yes/No & Num & Other & Other \\
			    \hline
			    UpDn~\cite{anderson2018bottom}$_{\textit{CVPR'18}}$ & UpDn &  &  & 39.74 & 42.27 & 11.93 & 46.05 & 63.48 & 81.18 & 42.14 & 55.66 & 9.61 \\
				AReg$^\dagger$~\cite{ramakrishnan2018overcoming}$_{\textit{NeurIPS'18}}$ & UpDn & & & 41.17 & 65.49 & 15.48 & 35.48 & 62.75 & 79.84 & 42.35 & 55.16 & 19.58 \\
				MuRel~\cite{cadene2019murel}$_{\textit{CVPR'19}}$ & UpDn & & & 39.54 & 42.85 & 13.17 & 45.04 & --- & --- & --- & --- &  --- \\
				GRL$^\dagger$~\cite{grand2019adversarial}$_{\textit{ACL'19}}$ & UpDn & & & 42.33 & 59.74 & 14.78 & 40.76 & 51.92 & --- & --- & --- & ---  \\
				RUBi$^\dagger$$^*$~\cite{cadene2019rubi}$_{\textit{NeurIPS'19}}$ & UpDn & & & 45.23 & 64.85 & 11.83 & 44.11 & 50.56 & 49.45 & 41.02 & 53.95 & 9.84  \\
				SCR~\cite{wu2019self}$_{\textit{NeurIPS'19}}$ & UpDn & & & 48.47 & 70.41 & 10.42 & 47.29 & 62.30 & 77.40 & 40.90 & 56.50 & 9.21 \\
				LMH$^\dagger$$^*$~\cite{clark2019don}$_{\textit{EMNLP'19}}$ & UpDn & & & 52.45 & 69.81 & 44.46 & 45.54 & 61.64 & 77.85 & 40.03 & 55.04 & 9.50 \\
				CVL~\cite{abbasnejad2020counterfactual}$_{\textit{CVPR'20}}$ & UpDn & & & 42.12 & 45.72 & 12.45 & 48.34 &  --- & --- & ---  & --- & --- \\
				Unshuffling~\cite{teney2020unshuffling}$_{\textit{arXiv'20}}$ & UpDn & & & 42.39 & 47.72 & 14.43 & 47.24 & 61.08 & 78.32 & 42.16 & 52.81 & 5.57 \\
				RandImg~\cite{teney2020value}$_{\textit{NeurIPS'20}}$ & UpDn & & & 55.37 & 83.89 & 41.60 & 44.20 & 57.24 & 76.53 & 33.87 & 48.57 & 4.37 \\
				SSL~\cite{zhu2020overcoming}$_{\textit{IJCAI'20}}$ & UpDn & & & 57.59 & 86.53 & 29.87 & 50.03 & 63.73 &  --- & --- & --- & --- \\
				CSS+CL$^\dagger$~\cite{liang2020learning}$_{\textit{EMNLP'20}}$ & UpDn & & & 59.18 & 86.99 & 49.89 & 47.16 & 57.29 & 67.27 & 38.40 & 54.71 & 7.55 \\
				CF-VQA$^\dagger$~\cite{niu2021counterfactual}$_{\textit{CVPR'21}}$ & UpDn & & & 53.55 & \textbf{91.15} & 13.03 & 44.97 & 63.54 & 82.51 & 43.96 & 54.30 & 9.33 \\
				GGE-DQ$^\dagger$~\cite{han2021greedy}$_{\textit{ICCV'21}}$ & UpDn & & & 57.32 & 87.04 & 27.75 & 49.59 & 59.11 & 73.27 & 39.99 & 54.39 & 4.80 \\
				LMH+SAR$^\dagger$$^*$~\cite{si2021check}$_{\textit{ACL'21}}$ & UpDn & \boldcheckmark & & 62.51 & 76.40 & \textbf{59.40} & 56.09 & 65.79 & 77.26 & 52.71 & 60.52 & 4.43\\
				\textbf{LMH-CSS}$_{\textit{CVPR'20}}$ & UpDn &  &  & \cellcolor{mygray-bg}{58.95} & \cellcolor{mygray-bg}{84.37} & \cellcolor{mygray-bg}{49.42} & \cellcolor{mygray-bg}{48.21} & \cellcolor{mygray-bg}{59.91} & \cellcolor{mygray-bg}{73.25} & \cellcolor{mygray-bg}{39.77} & \cellcolor{mygray-bg}{55.11} & \cellcolor{mygray-bg}{6.90} \\
				\textbf{LMH-CSS$^+$} & UpDn &  & & \cellcolor{mygray-bg}{59.54} & \cellcolor{mygray-bg}{83.37}  & \cellcolor{mygray-bg}{52.57}  & \cellcolor{mygray-bg}{48.97}  & \cellcolor{mygray-bg}{59.96} & \cellcolor{mygray-bg}{73.69}  & \cellcolor{mygray-bg}{40.18}  & \cellcolor{mygray-bg}{54.77}  & \cellcolor{mygray-bg}{5.80} \\
				\textbf{LMH-CSST} & UpDn &  & & \cellcolor{mygray-bg}{61.66} & \cellcolor{mygray-bg}{90.20}  & \cellcolor{mygray-bg}{54.42}  & \cellcolor{mygray-bg}{48.69}  & \cellcolor{mygray-bg}{62.37} & \cellcolor{mygray-bg}{80.05}  & \cellcolor{mygray-bg}{39.24}  & \cellcolor{mygray-bg}{55.04}  & \cellcolor{mygray-bg}{6.35} \\
				\textbf{LMH-CSST+SAR} & UpDn & \boldcheckmark & & \cellcolor{mygray-bg}{\textbf{66.49}} & \cellcolor{mygray-bg}{86.97}  & \cellcolor{mygray-bg}{57.95}  & \cellcolor{mygray-bg}{\textbf{58.53}} & \cellcolor{mygray-bg}{\textbf{69.31}} & \cellcolor{mygray-bg}{\textbf{85.85}} & \cellcolor{mygray-bg}{\textbf{52.87}} & \cellcolor{mygray-bg}{\textbf{61.08}} & \cellcolor{mygray-bg}{\textbf{2.55}} \\
				\hline
				UpDn$^{*\ddag}$~\cite{anderson2018bottom}$_{\textit{CVPR'18}}$ & LXMERT & \boldcheckmark & &  44.14 & 43.12  & 17.07 & 51.66  & 67.69 & 83.83 & 50.80 & 59.89 & 8.23 \\
				LMH$^{*\dagger\ddag}$~\cite{clark2019don}$_{\textit{EMNLP'19}}$ & LXMERT & \boldcheckmark & & 59.66 & 73.41 & 57.72 & 52.99 & 59.57 & 64.17  & 47.51 & 59.27 & 6.28 \\
			    \textbf{LMH-CSS}$^+$$^\ddag$ & LXMERT & \boldcheckmark & & \cellcolor{mygray-bg}{63.63} & \cellcolor{mygray-bg}{84.70}  & \cellcolor{mygray-bg}{62.12}  & \cellcolor{mygray-bg}{53.00}  & \cellcolor{mygray-bg}{58.01} & \cellcolor{mygray-bg}{60.37}  & \cellcolor{mygray-bg}{47.02}  & \cellcolor{mygray-bg}{59.14} & \cellcolor{mygray-bg}{6.14} \\
				\textbf{LMH-CSST}$^\ddag$ & LXMERT & \boldcheckmark & & \cellcolor{mygray-bg}{65.71} & \cellcolor{mygray-bg}{\textbf{90.10}}  & \cellcolor{mygray-bg}{\textbf{63.70}}  & \cellcolor{mygray-bg}{53.48} & \cellcolor{mygray-bg}{65.71} & \cellcolor{mygray-bg}{80.61}  & \cellcolor{mygray-bg}{48.26}  & \cellcolor{mygray-bg}{58.99} & \cellcolor{mygray-bg}{5.51} \\
				\textbf{LMH-CSST+SAR}$^\ddag$ & LXMERT & \boldcheckmark & & \cellcolor{mygray-bg}{\textbf{67.49}} & \cellcolor{mygray-bg}{88.06}  & \cellcolor{mygray-bg}{56.57}  & \cellcolor{mygray-bg}{\textbf{59.71}}  & \cellcolor{mygray-bg}{\textbf{69.32}} & \cellcolor{mygray-bg}{\textbf{84.11}}  & \cellcolor{mygray-bg}{\textbf{54.90}}  & \cellcolor{mygray-bg}{\textbf{61.88}} & \cellcolor{mygray-bg}{\textbf{2.17}} \\
				\hline
                MUTANT~\cite{gokhale2020mutant}$_{\textit{EMNLP'20}}$ & UpDn & & \boldcheckmark & 61.72 & 88.90 & 49.68 & 50.78 & 62.56 & 82.07 & 42.52 & 53.28 & 2.50 \\
				MUTANT~\cite{gokhale2020mutant}$_{\textit{EMNLP'20}}$ & LXMERT & \boldcheckmark & \boldcheckmark & 69.52 & 93.15 & 67.17 & 57.78 & 70.24 & 89.01 & 54.21 & 59.96 & 2.18 \\
				\hline
			\end{tabular}
		} 
	\end{center}
	\vspace{-2em}
	\label{tab:SOTA_v2}
\end{table*}

\addtolength{\tabcolsep}{-2pt}
\begin{table}[tbp]
	\small
	\caption{Accuracies (\%) on VQA-CP v1 test set of state-of-the-art models. Results are shown for ``all'' (All) questions, ``yes/no'' (Y/N) questions, ``number'' (Num) questions, and ``other'' (Other) questions. $^\dagger$ denotes ensemble-based methods. $^*$ indicates results from our reimplementation. $^\ddag$ denotes models use the adapting strategy for LXMERT backbone, and SAR~\cite{si2021check} denotes models use an extra ranking stage (cf. Sec.~\ref{sec:expsetting}).}
	\vspace{-1em}
	\begin{center}
		\scalebox{0.98}{
			\begin{tabular}{| l | c | c c c c|}
				\hline
				Models & Base & All & Y/N & Num & Other  \\
				\hline
				GVQA~\cite{agrawal2018don}$_{\textit{CVPR'18}}$ & SAN & 39.23 & 64.72 & 11.87 & 24.86 \\
				UpDn~\cite{anderson2018bottom}$_{\textit{CVPR'18}}$ & UpDn & 39.74 & 42.27 & 11.93 & 46.05 \\
				AReg$^\dagger$~\cite{ramakrishnan2018overcoming}$_{\textit{NeurIPS'18}}$ & UpDn & 41.17 & 65.49 & 15.48 & 35.48 \\
				GRL$^\dagger$~\cite{grand2019adversarial}$_{\textit{ACL'19}}$ & UpDn & 45.69 & 77.64 & 13.21 & 26.97 \\
				RUBi$^\dagger$$^*$~\cite{cadene2019rubi}$_{\textit{NeurIPS'19}}$ & UpDn & 50.90 & 80.83 & 13.84 & 36.02 \\
				LMH$^\dagger$$^*$~\cite{clark2019don}$_{\textit{EMNLP'19}}$ & UpDn & 55.27 & 76.47 & 26.66 & 45.68 \\
				\textbf{LMH-CSS}$_{\textit{CVPR'20}}$ & UpDn & 60.95 & 85.60 & 40.57 & 44.62  \\
				\textbf{LMH-CSS$^+$} & UpDn & \cellcolor{mygray-bg}{61.66} & \cellcolor{mygray-bg}{84.97} & \cellcolor{mygray-bg}{41.65}  & \cellcolor{mygray-bg}{46.51}  \\
				\textbf{LMH-CSST} & UpDn & \cellcolor{mygray-bg}{63.16} & \cellcolor{mygray-bg}{90.58} & \cellcolor{mygray-bg}{39.16}  & \cellcolor{mygray-bg}{45.53}  \\
				\textbf{LMH-CSST+SAR} & UpDn & \cellcolor{mygray-bg}{\textbf{69.57}} & \cellcolor{mygray-bg}{\textbf{90.81}}  & \cellcolor{mygray-bg}{\textbf{49.29}} & \cellcolor{mygray-bg}{\textbf{56.59}}   \\
				\hline
				\textbf{LMH-CSS$^+$}$^\ddag$ & LXMERT & \cellcolor{mygray-bg}{66.13} & \cellcolor{mygray-bg}{88.89} & \cellcolor{mygray-bg}{42.74}  & \cellcolor{mygray-bg}{52.88}  \\
				\textbf{LMH-CSST}$^\ddag$ & LXMERT & \cellcolor{mygray-bg}{67.27} & \cellcolor{mygray-bg}{92.04} & \cellcolor{mygray-bg}{43.23}  & \cellcolor{mygray-bg}{52.30}  \\
				\textbf{LMH-CSST+SAR}$^\ddag$ & LXMERT & \cellcolor{mygray-bg}{\textbf{69.75}} & \cellcolor{mygray-bg}{\textbf{92.27}} & \cellcolor{mygray-bg}{\textbf{44.05}}  & \cellcolor{mygray-bg}{\textbf{57.69}}  \\
				\hline
			\end{tabular}
		} 
	\end{center}
	\label{tab:SOTA_v1}
        \vspace{-1em}
\end{table}
\addtolength{\tabcolsep}{2pt}

\addtolength{\tabcolsep}{-2pt} 
\begin{table}[tbp]
	\small
	\caption{Accuracies (\%) on GQA-OOD testdev set of state-of-the-art models. Results are shown for ``all'' (All) questions, ``yes/no'' (Y/N) questions, ``number'' (Num) questions, and ``other'' (Other) questions. $^*$ indicates the results from our reimplementation using officially released codes. $^\ddag$ denotes models use the adapting strategy for LXMERT backbone, and SAR~\cite{si2021check} denotes models use an extra ranking stage (cf. Sec.~\ref{sec:expsetting}).}
	\vspace{-1em}
	\begin{center}
		\scalebox{0.98}{
			\begin{tabular}{| l | c | c c c c|}
				\hline
				Models & Base & Acc-A & Acc-T & Acc-H & $\Delta$$\downarrow$  \\
				\hline
				MCAN~\cite{yu2019deep}$_{\textit{CVPR'19}}$ & MCAN & 50.8 & 46.5 & 53.4 & 14.8 \\
				BAN4~\cite{kim2018bilinear}$_{\textit{NeurIPS'18}}$ & BAN4 & 50.2 & 47.2 & 51.9 & 9.9 \\
				
				UpDn$^*$$^\ddag$~\cite{anderson2018bottom}$_{\textit{CVPR'18}}$ & LXMERT & 49.61 & 46.10 & 51.76 & 12.3 \\
				LMH$^*$$^\ddag$~\cite{clark2019don}$_{\textit{EMNLP'19}}$ & LXMERT & 47.78 & 45.44 & 49.22 & \textbf{8.3} \\
                \textbf{LMH-CSS}$^\ddag$$_{\textit{CVPR'20}}$ & LXMERT & 48.10 & 44.59 & 50.26 & 12.7 \\
				\textbf{LMH-CSS$^+$}$^\ddag$ & LXMERT & \cellcolor{mygray-bg}{49.21} & \cellcolor{mygray-bg}{46.28} & \cellcolor{mygray-bg}{51.01}  & \cellcolor{mygray-bg}{10.2} \\
				\textbf{LMH-CSST}$^\ddag$ & LXMERT & \cellcolor{mygray-bg}{49.75} & \cellcolor{mygray-bg}{46.38}  & \cellcolor{mygray-bg}{51.82}  & \cellcolor{mygray-bg}{11.7} \\
				\textbf{LMH-CSST+SAR}$^\ddag$ & LXMERT & \cellcolor{mygray-bg}{\textbf{51.36}} & \cellcolor{mygray-bg}{\textbf{48.07}}  & \cellcolor{mygray-bg}{\textbf{53.38}}  & \cellcolor{mygray-bg}{11.0} \\
				\hline\hline
			    UpDn$^*$~\cite{anderson2018bottom}$_{\textit{CVPR'18}}$ & UpDn & 46.96 & 42.90 &  49.45 & 15.3 \\
				RUBi$^*$~\cite{cadene2019rubi}$_{\textit{NeurIPS'19}}$  & UpDn & 45.85 & 43.37 & 47.37 & 9.2 \\
				LMH$^*$~\cite{clark2019don}$_{\textit{EMNLP'19}}$ & UpDn  & 43.96 & 40.73 & 45.93 & 12.8 \\
                \textbf{LMH-CSS}$^\ddag$$_{\textit{CVPR'20}}$ & UpDn & 43.10 & 41.49 & 44.09 & \textbf{6.3} \\ 
				\textbf{LMH-CSS$^+$} & UpDn & \cellcolor{mygray-bg}{44.24} & \cellcolor{mygray-bg}{41.20}  & \cellcolor{mygray-bg}{46.11}  & \cellcolor{mygray-bg}{11.9} \\
				\textbf{LMH-CSST} & UpDn & \cellcolor{mygray-bg}{45.42} & \cellcolor{mygray-bg}{42.90}  & \cellcolor{mygray-bg}{46.97}  & \cellcolor{mygray-bg}{9.5}   \\
				\textbf{LMH-CSST+SAR} & UpDn & \cellcolor{mygray-bg}{\textbf{51.07}} & \cellcolor{mygray-bg}{\textbf{47.98}}  & \cellcolor{mygray-bg}{\textbf{52.97}}  & \cellcolor{mygray-bg}{10.4}  \\
				\hline
			\end{tabular}
		} 
	\end{center}
	\label{tab:SOTA_GQA}
        \vspace{-2em}
\end{table}
\addtolength{\tabcolsep}{2pt}

\subsection{Ablation Studies }
In this subsection, we mainly focus on the ablation studies to verify the robustness and generalization ability of our proposed CSST. Specifically, firstly, we conducted a set of ablations on different hyperparameters of CSS in Sec.~\ref{sec:4.3.1}. Secondly, we incorporated CSST into different VQA baselines to show its architecture generalization ability in Sec.~\ref{sec:4.3.2}. Third, we conducted ablation studies on the influence of positive and negative sample selection to show the importance of the sample selection in CST in Sec.~\ref{sec:4.3.3}.

\subsubsection{Influence of Different Hyperparameters of CSS} \label{sec:4.3.1}
We run a number of ablation studies to analyze the influence of different hyperparameters of CSS$^+$ (\ie, V-CSS$^+$ and Q-CSS$^+$\footnote{V-CSS$^+$ and Q-CSS$^+$ denote the improved version of V-CSS and Q-CSS, respectively (cf. Sec.~\ref{sec:3.2.3}).}). Specifically, we conducted all ablations by building on top of a typical ensemble-based VQA model LMH~\cite{clark2019don}. To disentangle the influence of our contrastive training, we only use the XE loss as training objective and both original samples and counterfactual samples as the inputs (cf. Fig.~\ref{fig:2} (b), similar with~\cite{chen2020counterfactual}). All results are illustrated in Fig.~\ref{fig:5}.

\noindent\textbf{Size of $\mathcal{I}$ in V-CSS$^+$.} The influence of different size of $\mathcal{I}$ is shown in Fig.~\ref{fig:5} (a). We can observe that the model's performance gradually decreases with the increase of $|\mathcal{I}|$. 

\noindent\textbf{Size of critical objects in V-CSS$^+$.} The influence of masking different numbers of critical objects is shown in Fig.~\ref{fig:5} (a). We compared dynamic K (cf. Eq.~\eqref{eq:topk_objects}) with some fixed constants (\eg, 1, 3, 5). From the results, we can observe that the model with dynamic K achieves the best performance.

\begin{table*}[!t]
    \centering
	\caption{Quantitative (\%) results about the evaluation of the VQA models' visual-explainable and question-sensitive abilities. Results are shown for ``all'' (All) questions, ``yes/no'' (Y/N) questions, ``number'' (Num) questions, and ``other'' (Other) questions.}
    \vspace{-0.5em}
	\subfloat[Accuracies (\%) on VQA-CP v2 test set.]{
		\tablestyle{2.5pt}{1.05}\begin{tabular}{l|x{25}x{25}x{25}x{25}}
			\hline
			Models & All & Y/N & Num & Other  \\
			\hline
			\multicolumn{5}{l}{\emph{Models with UpDn backbone}} \\
			\hline
			SCR & 48.47 & 70.41 & 10.42 & 47.29 \\
			LMH &52.45 & 69.81 & 44.46 & 45.54  \\
			~+SCR & \multicolumn{4}{c}{continued decrease} \\
            ~+CSS & 58.95 & 84.37 & 49.42 & 48.21 \\
 			~+\textbf{CSS$^+$} & \cellcolor{mygray-bg}{59.54} & \cellcolor{mygray-bg}{83.37}  & \cellcolor{mygray-bg}{52.57}  & \cellcolor{mygray-bg}{\textbf{48.97}} \\
			~+\textbf{CSST} & \cellcolor{mygray-bg}{\textbf{61.66}} & \cellcolor{mygray-bg}{\textbf{90.20}}  & \cellcolor{mygray-bg}{\textbf{54.42}}  & \cellcolor{mygray-bg}{48.69} \\
			\hline
			\multicolumn{5}{l}{\emph{Models with LXMERT backbone}} \\
			\hline
			LMH & 59.66 & 73.41 & 57.72 & 52.99 \\                
			~+CSS & 63.08 & 85.80 & 58.76 & 52.35 \\ 
			~+\textbf{CSS$^+$} & \cellcolor{mygray-bg}{63.63} & \cellcolor{mygray-bg}{84.70}  & \cellcolor{mygray-bg}{62.12}  & \cellcolor{mygray-bg}{53.00} \\
			~+\textbf{CSST} & \cellcolor{mygray-bg}{\textbf{65.71}} & \cellcolor{mygray-bg}{\textbf{90.10}}  & \cellcolor{mygray-bg}{\textbf{63.70}}  & \cellcolor{mygray-bg}{\textbf{53.48}} \\
			\hline
	\end{tabular}}\hspace{3mm}
	\subfloat[$\mathcal{AI}$ score (\%) on VQA-CP v2 test set.]{
		\tablestyle{2.5pt}{1.05}\begin{tabular}{l|x{25}x{25}x{25}x{25}}
			\hline
			Models & Top-1 & Top-2 & Top-3  \\
			\hline
			\multicolumn{4}{l}{\emph{Models with UpDn backbone}} \\
			\hline
			SCR & 15.55 & 13.40 & 11.97 \\
			LMH & 17.81 & 15.19 & 13.52 \\
            ~+V-CSS & 18.14 & 14.91 & 13.06 \\ 
			~+V-CSS$^+$ & \cellcolor{mygray-bg}{\textbf{20.66}} & \cellcolor{mygray-bg}{\textbf{16.71}}  & \cellcolor{mygray-bg}{14.34} \\
            ~+CSS & 16.54 & 14.08 & 12.62 \\
			 ~+\textbf{CSS$^+$} & \cellcolor{mygray-bg}{18.81} & \cellcolor{mygray-bg}{15.73}  & \cellcolor{mygray-bg}{13.84} \\
			~+\textbf{CSST} & \cellcolor{mygray-bg}{19.54} & \cellcolor{mygray-bg}{16.37}  & \cellcolor{mygray-bg}{\textbf{14.41}} \\
			\hline
			\multicolumn{4}{l}{\emph{Models with LXMERT backbone}} \\
			\hline			
			LMH & 17.16 & 15.06 & 13.60 \\
            ~+CSS & 18.71 & 16.15 & 14.35 \\ 
			~+\textbf{CSS$^+$} & \cellcolor{mygray-bg}{18.90} & \cellcolor{mygray-bg}{16.28}  & \cellcolor{mygray-bg}{14.43} \\
			~+\textbf{CSST} & \cellcolor{mygray-bg}{\textbf{22.68}} & \cellcolor{mygray-bg}{\textbf{18.99}}  & \cellcolor{mygray-bg}{\textbf{16.48}} \\
			\hline
	\end{tabular}}\hspace{3mm}
	\subfloat[\textbf{Left}: $CS(k)$ (\%) on VQA-CP-Rephrasing; \textbf{Right}: $\mathcal{CI}$ score (\%) on VQA-CP v2 test set.]{
		\tablestyle{2.5pt}{1.05}\begin{tabular}{l|x{22}x{22}x{22}x{22}|x{22}}
			\hline
			Models & K=1 & K=2 & K=3 & K=4  & $\mathcal{CI}$ \\
			\hline
			\multicolumn{6}{l}{\emph{Models with UpDn backbone}} \\
			\hline
			UpDn & 49.94 & 38.80 & 31.55 & 28.08 & 33.70 \\
			LMH & 51.68 & 39.84 & 33.38 & 29.11 & 40.26 \\
            ~+Q-CSS & 55.50 & 43.79 & 37.09 & 32.60 & 55.76 \\ 
			~+Q-CSS$^+$ & \cellcolor{mygray-bg}{55.69} & \cellcolor{mygray-bg}{43.95}  & \cellcolor{mygray-bg}{37.22}  & \cellcolor{mygray-bg}{32.68} & \cellcolor{mygray-bg}{57.08}  \\
			~+CSS & 54.98 & 42.09 & 34.74 & 29.98 & 54.28 \\
			~+\textbf{CSS$^+$} & \cellcolor{mygray-bg}{56.31} & \cellcolor{mygray-bg}{44.08}  & \cellcolor{mygray-bg}{37.00}  & \cellcolor{mygray-bg}{32.29} & \cellcolor{mygray-bg}{51.87} \\
			~+\textbf{CSST} & \cellcolor{mygray-bg}{\textbf{59.97}} & \cellcolor{mygray-bg}{\textbf{49.49}}  & \cellcolor{mygray-bg}{\textbf{43.37}}  & \cellcolor{mygray-bg}{\textbf{39.24}} & \cellcolor{mygray-bg}{\textbf{58.45}} \\
			\hline
			\multicolumn{6}{l}{\emph{Models with LXMERT backbone}} \\
			\hline
			LMH & 61.21 & 50.33 & 43.94 & 39.52 & 55.42 \\
            ~+CSS & 61.68 & 50.94 & 44.60 & 40.23 & 55.18 \\ 
			~+\textbf{CSS$^+$} & \cellcolor{mygray-bg}{61.75} & \cellcolor{mygray-bg}{50.62}  & \cellcolor{mygray-bg}{43.92}  & \cellcolor{mygray-bg}{39.27} & \cellcolor{mygray-bg}{55.44} \\
			~+\textbf{CSST} & \cellcolor{mygray-bg}{\textbf{63.87}} & \cellcolor{mygray-bg}{\textbf{53.72}}  & \cellcolor{mygray-bg}{\textbf{47.70}}  & \cellcolor{mygray-bg}{\textbf{43.58}} & \cellcolor{mygray-bg}{\textbf{58.22}} \\
			\hline
	\end{tabular}}\hspace{3mm}
	\vspace{-1em}
	\label{tab:vq_ablatives}
\end{table*}

\noindent\textbf{Size of critical words in Q-CSS$^+$.} The influence of replacing different sizes of critical words is shown in Fig.~\ref{fig:5} (b). From the results, we can observe that replacing only one word (\ie, top-1) achieves the best performance. Of course, we can also use the same dynamic way as V-CSS$^+$ to choose the top-k critical words in Q-CSS$^+$. However, in our earlier exploration experiments, we found that the dynamic method achieves slightly worse results than these fixed settings.

\noindent\textbf{Proportion $\delta$ between V-CSS$^+$ and Q-CSS$^+$.} The influence of different $\delta$ is shown in Fig.~\ref{fig:5} (c). From the results, we can observe that the performance is best when $\delta = 0.5$.

In all the following experiments (including experiments on different benchmarks, models with different backbones, and models with contrastive training), we used the same best hyperparameter settings for both V-CSS$^+$ and Q-CSS$^+$.

\subsubsection{Architecture Generalization of CSST} \label{sec:4.3.2}

\noindent\textbf{Settings.} Since our CSST is a model-agnostic training strategy, which can be seamlessly incorporated into any different VQA architecture. To evaluate the effectiveness of CSST to boost the debiasing performance of different backbones, we incorporated the CSST into multiple architectures including: UpDn~\cite{anderson2018bottom}, RUBi~\cite{cadene2019rubi}, LMH~\cite{clark2019don}. Especially, RUBi and LMH are ensemble-based methods. For VQA-CP v2, we followed the same settings as prior works. For GQA-OOD, we used the adapting strategies mentioned in Sec.~\ref{sec:expsetting}. All results are shown in TABLE~\ref{tab:boost_models}. For more clear comparisons, we used superscript $^+$ to distinguish this improved version of CSS (Q-CSS/V-CSS) from their respective initial counterparts~\cite{chen2020counterfactual}.

\noindent\textbf{Results.} Compared to these baseline models, our CSST (\ie, both CSST-G and CSST-L) can consistently improve the performance for all architectures. For different architectures, the behaviors of CSST-G and CSST-L are slightly different: CSST-G is more suitable for complex ensemble-based models (\eg, 9.21\% absolute gains in LMH on VQA-CP v2), and CSST-L is more suitable for vanilla VQA models (\eg, 16.87\% absolute gains in UpDN on VQA-CP v2). Meanwhile, when both two types of CSS are used (\ie, Q-CSS$^+$ and V-CSS$^+$), models often achieve better performance than single type CSS. Compared to CSS~\cite{chen2020counterfactual}, the new CSS$^+$ achieves better performance on most datasets and baselines (\eg, RUBi/LMH on VQA-CP v2 and GQA-OOD). Furthermore, the model with CSST (\ie, CSST-G or CSST-L) always achieves the best performance.

\subsubsection{Influence of Positive \& Negative Sample Selection} \label{sec:4.3.3}
 
\noindent\textbf{Settings.} To demonstrate the effectiveness of our proposed sample selection strategy, we compared CSST with a strong baseline CL~\cite{liang2020learning}, which has shown effectiveness under the CSS mechanism. Since the contributions of our CR are orthogonal to CL, we further reported the results by incorporating CR with CL, denoted as ``CL+CR''. Meanwhile, to further show the importance of \emph{counterfactual samples} in the sample selection, we separate these four negative samples into two groups: 1) ``CSS'': the first two negative samples from V-CSS$^+$ and Q-CSS$^+$; 2) ``Random'' (Rand): the last two randomly composed negative samples (cf. \textsc{NEG\_Sel} in Algorithm~\ref{alg:cst}). All results are reported in TABLE~\ref{tab:pos&neg}.

\noindent\nonumber\textbf{Results.} Compared to the CL baseline, model with CSST can achieve much better performance (\eg, 61.66\% vs. 59.78\%), which shows the importance of training sample diversity in contrastive training. However, the performance of CL+CR is even slightly worse than CR (61.36\% in CL+CR vs. 61.66\% in CR), \ie, overmuch these naive positive/negative samples from CL may even harm the diversity of samples and model training. Compared to the model with ``Rand'' negative samples, model with ``CSS'' negative samples also achieves better performance (\eg, 61.13\% vs. 60.74\%), which shows the importance of fine-grained differences in the samples for contrastive training. Meanwhile, when all negative samples are used, the model achieves the best performance.

\begin{figure*}[htbp]
    \centering
    \includegraphics[width=0.85\linewidth]{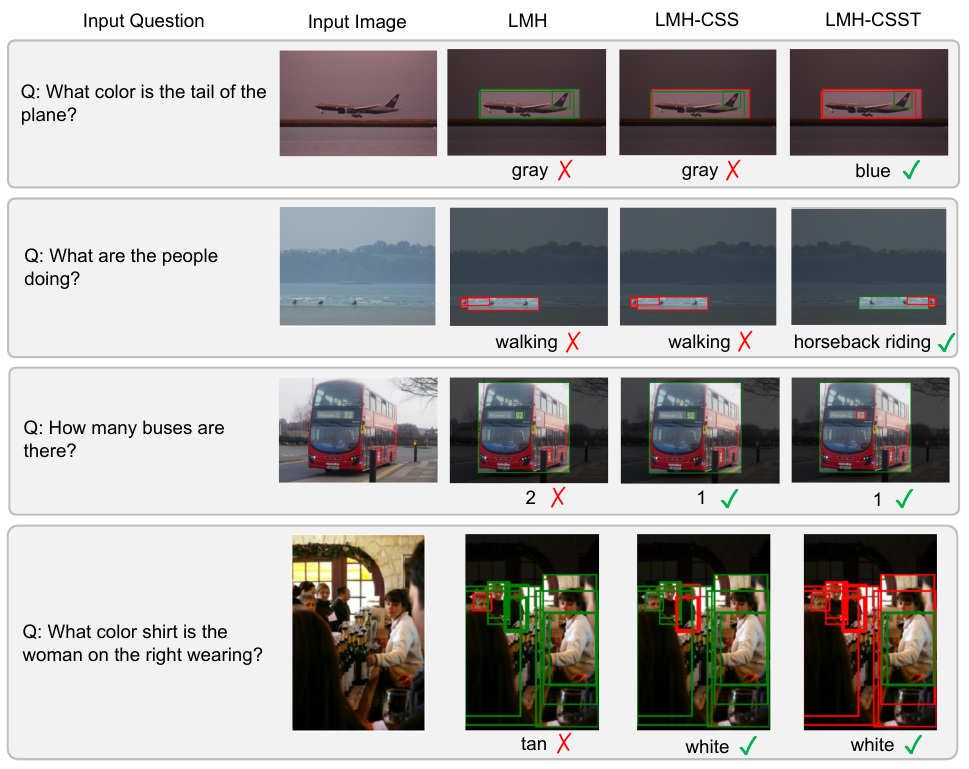}
    \vspace{-1em}
    \caption{Examples of visual-explainable ability from VQA-CP v2 test set. The \textcolor{green}{\textbf{green}} boxes denote their scores $s(\hat{a}, \bm{v})$ \textgreater $0$, \ie, positive contributions to final predictions. The \textcolor{red}{\textbf{red}} boxes denote their scores $s(\hat{a}, \bm{v})$ \textless $0$, \ie, negative contributions to final predictions. Only objects which are highly related to the QA pair are shown (\ie, $\mathcal{SIM} \geq 0.6$). LMH-CSS is the model with only XE loss~\cite{chen2020counterfactual}}
    \label{fig:6}
\end{figure*}

\begin{figure*}[htbp]
    \centering
    \includegraphics[width=0.85\linewidth]{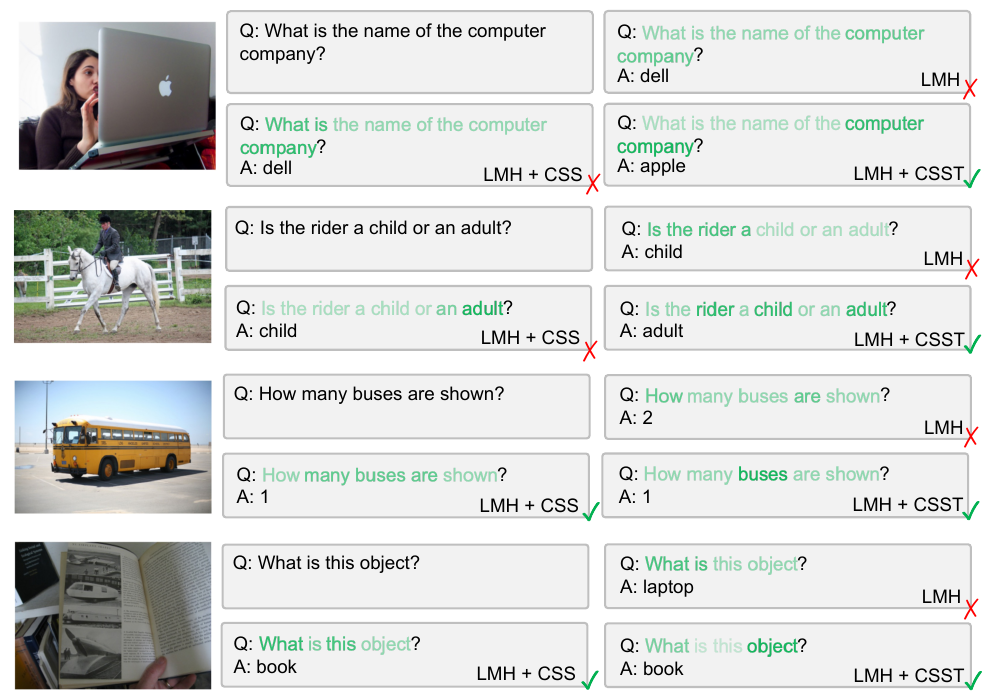}
    \vspace{-1.0em}
    \caption{Visualization examples of question-sensitive ability from VQA-CP v2 test set. Different shades of green color in the question denote the relative values of $s(\hat{a}, \bm{w})$, \ie, the word with darker green denotes the word has larger contributions to final predictions.}
    \label{fig:7}
     \vspace{-1em}
\end{figure*}

\begin{figure*}[t]
    \centering
    \includegraphics[width=0.85\linewidth]{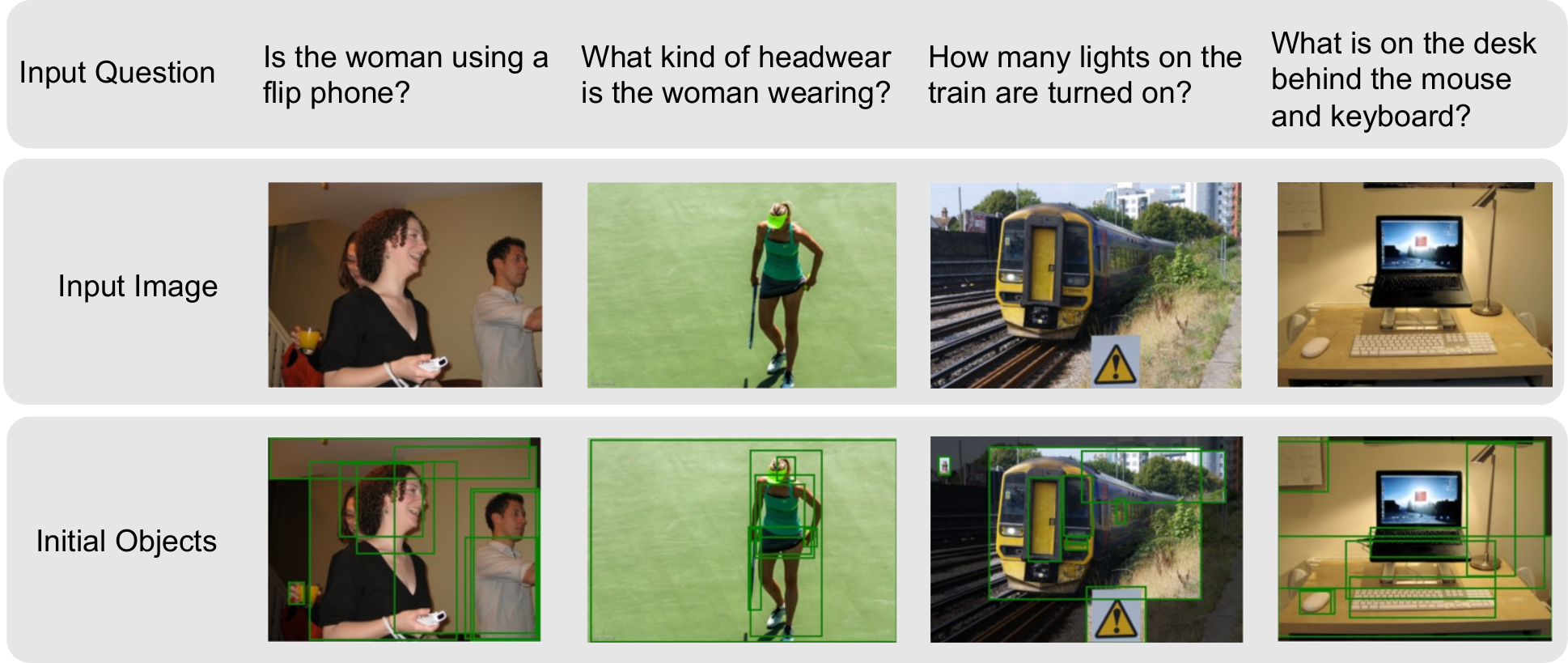}
    \vspace{-0.5em}
    \caption{Failure examples about \textbf{initial objects selection (\textsc{IO\_Sel} in Algorithm~\ref{alg:css})} of V-CSS. Samples are from the VQA-CP v2 test set. The green boxes denote the selected top-9 proposals for the initial object set.}
    \label{fig:failure_case1}
\end{figure*}

\subsection{Comparisons with State-of-the-Arts Models} \label{sec:SOTA}

In this subsection, we incorporated the CSST-G into model LMH~\cite{clark2019don}, which is dubbed as \textbf{LMH-CSST}. Then, we compared LMH-CSST with state-of-the-art methods.

\subsubsection{Performance on VQA-CP v2 and VQA v2} \label{sec:4.4.1}
\noindent\textbf{Settings.} We compared LMH-CSST with the state-of-the-art models on both VQA-CP v2 and VQA v2. According to the model framework, we group them into: 1) AReg~\cite{ramakrishnan2018overcoming}, GRL~\cite{grand2019adversarial}, RUBi~\cite{cadene2019rubi}, LMH~\cite{clark2019don}, CSS+CL~\cite{liang2020learning}, CF-VQA~\cite{niu2021counterfactual}, GGE-DQ~\cite{han2021greedy}, and LMH+SAR~\cite{si2021check}. They are all ensemble-based models. 2) UpDn~\cite{anderson2018bottom}, MuRel~\cite{cadene2019murel}, SCR~\cite{wu2019self}, CVL~\cite{abbasnejad2020counterfactual}, Unshuffling~\cite{teney2020unshuffling}, RandImg~\cite{teney2020value}, and SSL~\cite{zhu2020overcoming}. These models are vanilla VQA models. For complete comparisons, we reported results on two backbones: UpDn~\cite{anderson2018bottom} and LXMERT~\cite{tan2019lxmert}. For the UpDn backbone, we followed the same settings as prior works~\cite{clark2019don}. For the LXMERT backbone, we used adapting strategies mentioned in Sec.~\ref{sec:expsetting}. Since SAR~\cite{si2021check} is another model-agnostic model, which is orthogonal to our contributions, \ie, we can also equip LMH-CSST with SAR (dubbed as \textbf{LMH-CSST+SAR}) to further boost performance. Moreover, following suggestions from~\cite{teney2020value}, we also reported the performance gap in the \emph{Other} category.

\noindent\textbf{Results.} The results are reported in TABLE~\ref{tab:SOTA_v2}. When trained and tested on the VQA-CP v2 dataset (\ie, the left side of TABLE~\ref{tab:SOTA_v2}), LMH-CSST+SAR achieves a new SOTA performance on both UpDn and LXMERT backbones, with 66.49\% and 67.49\% accuracies, respectively. Compared to all models without extra pretraining or annotations (\ie, ``Pre.'' and ``Ann.'' in TABLE~\ref{tab:SOTA_v2}), LMH-CSST still achieves SOTA performance with 61.66\% accuracies. Particularly, CSST improves the performance of LMH with 9.21\% (61.66\% vs. 52.45\%) and 6.05\% (65.71\% vs. 59.66\%) absolute gains on UpDn and LXMERT backbones, respectively. When trained and tested on the VQA v2 dataset (\ie, the right side of TABLE~\ref{tab:SOTA_v2}), CSST consistently improves the performance of LMH with 0.73\% (62.37\% vs. 61.64\%) and 6.14\% (65.71\% vs. 59.57\%) absolute gains on UnDn and LXMERT backbones, respectively. Different from previous SOTA models that suffer severe performance drops between VQA-CP v2 and VQA v2 (\eg, 9.50\% and 6.28\% on LMH), LMH-CSST+SAR can significantly decrease the performance gap into 2.55\% and 2.17\%, which demonstrates that the effectiveness of CSST to reduce language biases.

\subsubsection{Performance on VQA-CP v1} \label{sec:4.4.2}

\noindent\textbf{Settings.} We also compared the LMH-CSST with state-of-the-art models on the VQA-CP v1. Similarly, we group these models into: 1) AReg~\cite{ramakrishnan2018overcoming}, GRL~\cite{grand2019adversarial}, RUBi~\cite{cadene2019rubi} and LMH~\cite{clark2019don} are all ensemble-based models. 2) GVQA~\cite{agrawal2018don} and UpDn are vanilla VQA models, and GVQA is with SAN~\cite{yang2016stacked} backbone. All settings are same as the ones on VQA-CP v2.

\noindent\textbf{Results.} The results are reported in TABLE~\ref{tab:SOTA_v1}. Compared to baseline models, both LMH-CSST and LMH-CSST+SAR achieve new state-of-the-art performance on two different backbones over all metrics. Particularly, the CSST improves the performance of LMH with a 7.89\% (63.16\% vs. 55.27\%) absolution performance gains on UpDn backbones.

\subsubsection{Performance on GQA-OOD} \label{sec:4.4.3}
\noindent\textbf{Settings.} We further compared LMH-CSST with state-of-the-art models on GQA-OOD. Similarly, we group these models into 1) RUBi~\cite{cadene2019rubi} and LMH~\cite{clark2019don} are ensemble-based models. 2) MCAN~\cite{yu2019deep}, BAN4~\cite{kim2018bilinear}, and UpDn are vanilla VQA models. All settings are same as experiments on VQA-CP. To adapt these ensemble-based methods to GQA-OOD, we used the adapting strategy mentioned in Sec.~\ref{sec:expsetting}.

\noindent\textbf{Results.} The results are reported in TABLE~\ref{tab:SOTA_GQA}. For fair comparisons, the baselines (UpDn and LMH) with the LXMERT backbones were reimplemented by using the same adapting strategy. Compared to baselines, LMH-CSST+SAR achieves new state-of-the-art performance on most of the metrics on both two backbones. Especially for the most important metrics Acc-T~\cite{kervadec2021roses}, our method achieves the best performance over others (\eg, 47.98\% vs. 43.37\% and 48.07\% vs. 47.2\%).

\subsection{Improving Visual-Explainable Ability}

We will validate the effectiveness of CSST to improve visual-explainable ability by answering the following questions: \textbf{Q1}: Can existing visual-explainable models be incorporated into the ensemble-based framework? \textbf{Q2}: How does CSST improve the model's visual-explainable ability?

\subsubsection{CSST vs. SCR (Q1)}

\noindent\textbf{Settings.} We equipped the existing state-of-the-art visual-explainable model SCR~\cite{wu2019self} into the LMH framework, and compared it with CSST. Results are reported in TABLE~\ref{tab:vq_ablatives} (a).

\noindent\textbf{Results.} Since the training of all SOTA visual-explainable models (\eg, SCR~\cite{wu2019self}, HINT~\cite{selvaraju2019taking}) are not end-to-end, for fair comparisons, we used a well-trained LMH (\ie, 52.45\% accuracies on VQA-CP v2) as the initial model. However, we observe that its performance continues to decrease from the start, which shows that the existing visual-explainable models can not be easily incorporated into the ensemble-based framework. In contrast, the proposed CSST can consistently improve performance on different backbones.

\subsubsection{Evaluations of Visual-Explainable Ability (Q2)}

\noindent\textbf{Settings.} We evaluated the effectiveness of CSS to improve the visual-explainable ability on both quantitative and qualitative results. For quantitative results, since we lack human annotations about the critical objects for each question, we regard the $\mathcal{SIM}$ scores (cf. \textsc{IO\_Sel} in Sec.~\ref{sec:v-css}) as pseudo ground-truths. Thus, we design a new metric \emph{Average Importance} ($\mathcal{AI}$): the average $\mathcal{SIM}$ score of the top-K objects with highest $|s(a, \bm{v})|$. We formally define $\mathcal{AI}$ as:
\begin{equation} 
\label{eq:AI}
\small
\mathcal{AI} = \frac{\sum_{(I, Q)} \left[ \mathbf{1}(a = \hat{a}) \cdot \sum_{k}\mathcal{SIM}_k^{(I, Q)} \right]}{\sum_{(I, Q)} 1},
\end{equation}
where $k$ is the index of top-K objects, $\mathcal{SIM}_k^{(I, Q)}$ is the $\mathcal{SIM}$ score of k-th object for the sample $(I, Q)$, and $\mathbf{1}$ is an indicator function. The results are shown in TABLE~\ref{tab:vq_ablatives} (b). We further demonstrate some qualitative results in Fig.~\ref{fig:6}.

\noindent\textbf{Results.} From TABLE~\ref{tab:vq_ablatives} (b), we can observe that CSST dramatically improves $\mathcal{AI}$ scores, which means the influential objects for predictions are more related to the QA pairs. From Fig.~\ref{fig:6}, we can find that both CSS (LMH vs. LMH-CSS) and CST (LMH-CSS vs. LMH-CSST) help the model to make predictions based on critical objects (\ie, green boxes) and suppress the influence of irrelevant objects (red boxes).

\subsection{Improving Question-Sensitive Ability}
We will validate the effectiveness of our CSST to improve the question-sensitive ability by answering the following questions: \textbf{Q3}: Does CSST help to improve the robustness to diverse rephrasings of questions? \textbf{Q4}: How does CSST improve the model's question-sensitive abilities?

\subsubsection{Robustness to Rephrasings of Questions (Q3)}
\noindent\textbf{Settings.} As discussed in previous work~\cite{shah2019cycle}, being robust to diverse rephrasing of questions is one of key behaviors of a question-sensitive model. To more accurately evaluate the robustness, we re-split the existing dataset VQA-Rephrasings~\cite{shah2019cycle} with the same splits as VQA-CP, and denoted it as VQA-CP-Rephrasings. For evaluation, we used the standard metric \emph{Consensus Score} $CS(k)$. Results are reported in TABLE~\ref{tab:vq_ablatives} (c) (left). We refer readers to~\cite{shah2019cycle} for more details about the VQA-Rephrasings and metric $CS(k)$.

\noindent\textbf{Results.} From TABLE~\ref{tab:vq_ablatives} (c), we can observe that Q-CSS$^+$ dramatically improves the robustness to diverse rephrasings of questions. Furthermore, V-CSS$^+$ can help to further improve the robustness, \ie, CSS$^+$ achieves better performance.

\begin{figure*}[t]
    \centering
    \includegraphics[width=0.85\linewidth]{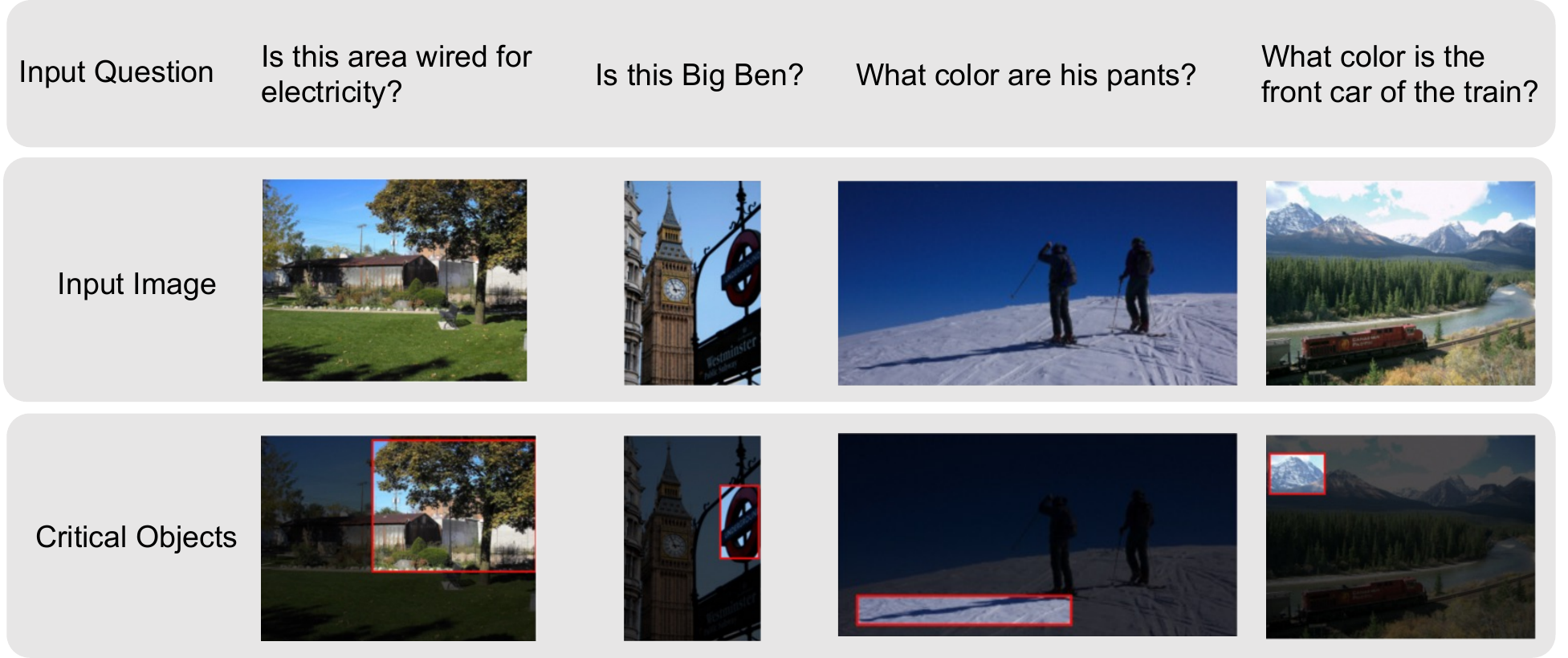}
    \vspace{-0.5em}
    \caption{Failure examples about \textbf{critical objects selection (\textsc{CO\_Sel} in Algorithm~\ref{alg:css})} of V-CSS. Samples are from the VQA-CP v2 test set. The red boxes denote the proposal with the highest Grad-CAM contribution for the ground-truth answer.}
    \vspace{-1em}
    \label{fig:failure_case2}
\end{figure*}

\begin{figure*}[h]
    \centering
    \includegraphics[width=0.95\linewidth]{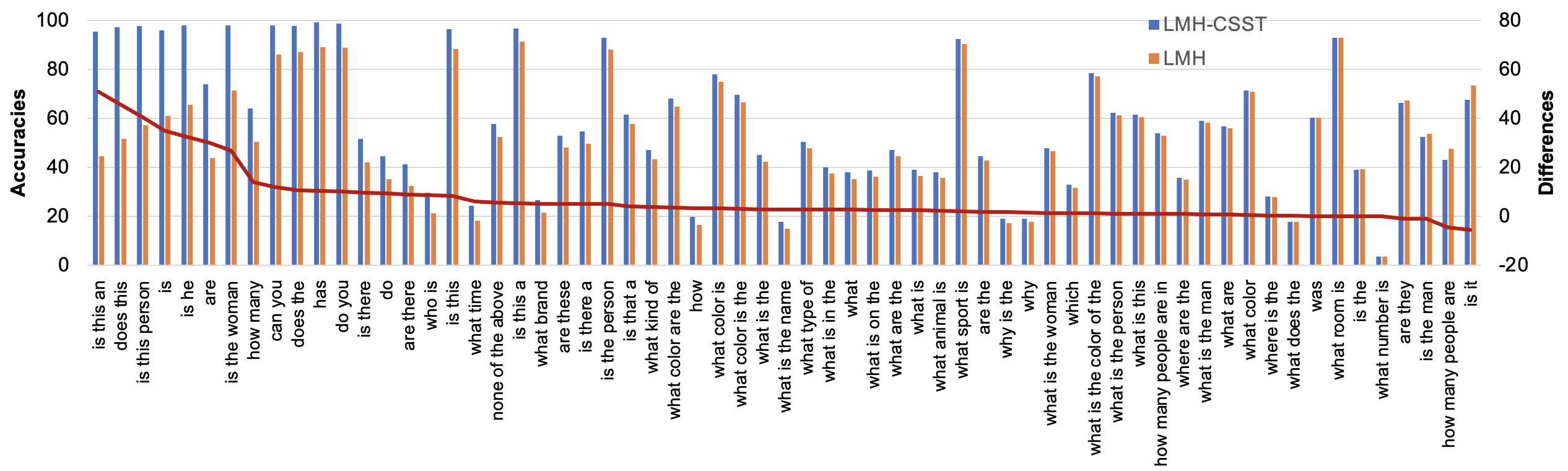}
    \vspace{-1em}
    \caption{\textbf{Left}: The VQA accuracy (\%) results of model LMH-CSST (blue bar) and LMH baseline (orange bar) on each different question type on VQA-CP v2 test set. The order is ranked by the performance improvements of our CSST on different question types. \textbf{Right}: The performance differences (\%) on each separate question type between LMH-CSST and baseline LMH.}
    \vspace{-1em}
    \label{fig:question-type}
\end{figure*}

\begin{table}[t]
    \caption{Accuracy (\%) of these question types that LMH-CSST performs worsen than LMH, \ie, Acc$_{\text{LMH}}$ $>$ Acc$_{\text{LMH-CSST}}$.}
    \vspace{-1.5em}
    \begin{center}
    \scalebox{0.95}{
        \begin{tabular}{| c | c c |}
            \hline
            Question Type & Acc$_{\text{LMH}}$ & Acc$_{\text{LMH-CSST}}$  \\
            \hline
                is the & 39.13  &  38.97 \\
                \textbf{how many people are} &  47.45  &  42.91\\
                \textbf{is it} & 73.33 & 67.66 \\
                are they & 67.33 & 66.22 \\
                is the man & 53.60 & 52.44 \\
                what number is & 3.54 & 3.34 \\
            \hline
        \end{tabular}
    }
    \end{center}
    \vspace{-2em}
    \label{tab:fail_types}
\end{table}

\subsubsection{Evaluations of Question-Sensitive Ability (Q4)}

\noindent\textbf{Settings.} We evaluated the effectiveness of CSST to improve the question-sensitive ability on quantitative and qualitative results. For quantitative results, since there is no standard evaluation metric, we design a new metric \emph{Confidence Improvement} ($\mathcal{CI}$): Given a test sample $(I, Q, a)$, we remove a critical noun in question $Q$, and obtain a new test sample $(I, Q^*, a)$\footnote{We directly masked the first noun of each question, and the noun is automatically detected by the POS tags using the spaCy POS tagger. The test set is released at: \href{https://github.com/yanxinzju/CSS-VQA}{https://github.com/yanxinzju/CSS-VQA}.}. Then we fed both two samples into the evaluated model, and calculated the confidence decreases of the ground-truth answer. We formally define $\mathcal{CI}$ as:
\begin{equation}
\small
\mathcal{CI} = \frac{\sum_{(I, Q)}  \mathbf{1}(P_{vqa}(a | I, Q) > P_{vqa}(a | I, Q^*)) \cdot \mathbf{1}(a = \hat{a}) }{\sum_{(I, Q)} 1}
\end{equation}
where $\hat{a}$ is the predicted answer for sample $(I, Q)$, $\mathbf{1}$ is an indicator function. The results are reported in TABLE~\ref{tab:vq_ablatives} (c). We further demonstrate some qualitative results in Fig.~\ref{fig:7}.

\noindent\textbf{Results.} From TABLE~\ref{tab:vq_ablatives} (c), we can observe that CSST helps the model to benefit more from the critical words, \ie, removing critical words results in more confidence drops for the ground-truth answers. From Fig.~\ref{fig:7}, we can find that both CSS (LMH vs. LMH-CSS) and CST (LMH-CSS vs. LMH-CSST) help the model to make predictions based on critical words (\eg, ``
bus'' or ``objects''), \ie, forcing the model to understand the whole question before making predictions.

\subsection{Failure Cases in CSST}

\noindent\textbf{Visualization of Failure Cases in V-CSS.}
We further illustrated some visualization examples of the failure cases on both Fig.~\ref{fig:failure_case1} and Fig.~\ref{fig:failure_case2}. Specifically, the errors in the V-CSS mainly come from two aspects: wrong initial object selection (cf. Fig.~\ref{fig:failure_case1}) and wrong critical object selection (cf. Fig.~\ref{fig:failure_case2}). For the initial object selection, as shown in Fig.~\ref{fig:failure_case1}, the initial object set is calculated by the cosine similarity between the detected object categories and nouns in the QA, thus, the initial object set may contain some irrelevant objects (\eg, ``\texttt{man}'' or ``\texttt{traffic sign}''). As for the critical objects, we visualized the proposal with the highest Grad-CAM contribution for the ground-truth answer. As shown in Fig.~\ref{fig:failure_case2}, the object with the highest Grad-CAM contributions may not be the critical object for the sample.

\noindent\textbf{Failure Question Patterns in CSST.} To further show the failure patterns of our CSST method, we reported the VQA accuracies over all the 65 question types on the VQA-CP v2 test set. The results are reported in Fig.~\ref{fig:question-type}. From Fig.~\ref{fig:question-type}, we can observe that in most of question types, CSST achieves better performance than the baseline model. Particularly, LMH-CSST surpasses LMH on all ``Other'' category questions, and only achieves worse results on 6 question types. We further listed the detailed performance of these types in TABLE~\ref{tab:fail_types}. From the table, we can find that CSST mainly doesn't help on some ``Yes/No'' and ``Number'' questions, such as the questions starting with ``\texttt{how many people are}'' or ``\texttt{is it}. As for the main reasons, just as discussed in the previous work~\cite{teney2020value}, it is more reliable to evaluate the results on the ``other'' categories, and the high performance (accuracy) on ``Y/N'' and ``Num'' questions can be easily obtained without any reasoning over text or images by simply exploiting knowledge about the dataset (\eg, LMH).

\section{Conclusions and Future Works}
In this paper, we proposed a model-agnostic Counterfactual Samples Synthesizing and Training (CSST) strategy to improve the VQA model's visual-explainable and question-sensitive abilities. The CSST is composed of two main components: CSS and CST, where CSS generates counterfactual training samples by masking critical objects or words, and CST consists of a XE training loss and a contrastive training loss. The CSST (\ie, CSS and CST) can consistently boost the performance of different VQA models. We validate the effectiveness of CSST through extensive comparative and ablative studies on three benchmarks and multiple different backbones. Moving forward, we are going to 1) extend CSST to other visual-language tasks that suffer severe language biases or other types of biases, \eg, captioning~\cite{chen2017sca}, grounding~\cite{chen2020rethinking}; 2) design better backbones that benefits from CSST.

\section*{Acknowledgment}
This work was supported by the National Key Research and Development Project of China (2021ZD0110700), the National Natural Science Foundation of China (U19B2043, 61976185), and the Fundamental Research Funds for the Central Universities (226-2022-00051). LC was supported by HKUST Special Support for Young Faculty (F0927).


%





\ifCLASSOPTIONcaptionsoff
  \newpage
\fi



\bibliographystyle{IEEEtran}
\bibliography{IEEEabrv,tpami21}
%



%




\vspace{-4em}

\begin{IEEEbiography}[{\includegraphics[width=1in,height=1.25in,clip,keepaspectratio]{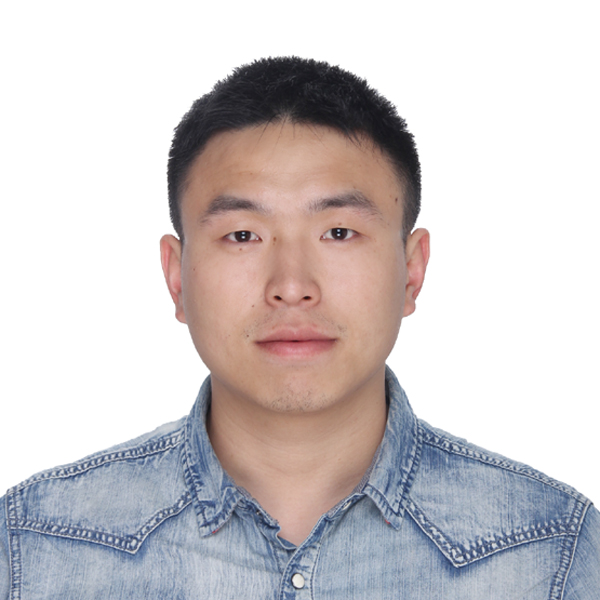}}]{Long Chen} received the Ph.D. degree in Computer Science from Zhejiang University in 2020, and the B.Eng. degree in Electrical Information Engineering from Dalian University of Technology in 2015. He is currently an assistant professor at The Hong Kong University of Science and Technology. He was a postdoctoral research scientist at Columbia University. 
His research interests are computer vision and multimedia. 
\end{IEEEbiography}

\vspace{-4em}

\begin{IEEEbiography}[{\includegraphics[width=1in,height=1.25in,clip,keepaspectratio]{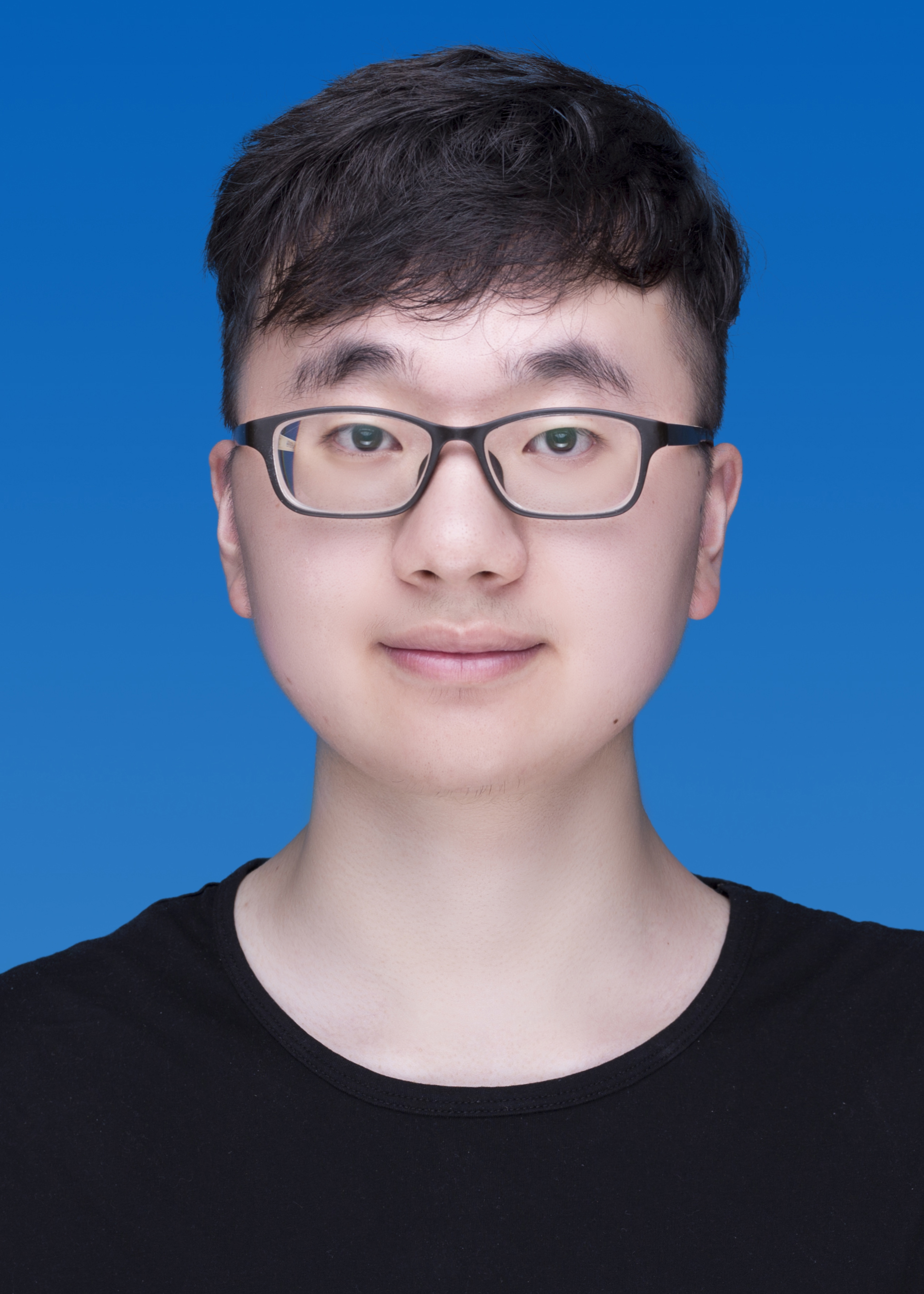}}]{Yuhang Zheng} received the B.Eng. degree in Computer Science from Zhejiang University, Hangzhou, China, in 2020. He is currently working toward his MS. degree in Computer Science at Zhejiang University, Hangzhou, China. His current research focuses on computer vision and multimedia.
\end{IEEEbiography}

\vspace{-4em}

\begin{IEEEbiography}[{\includegraphics[width=1in,height=1.25in,clip,keepaspectratio]{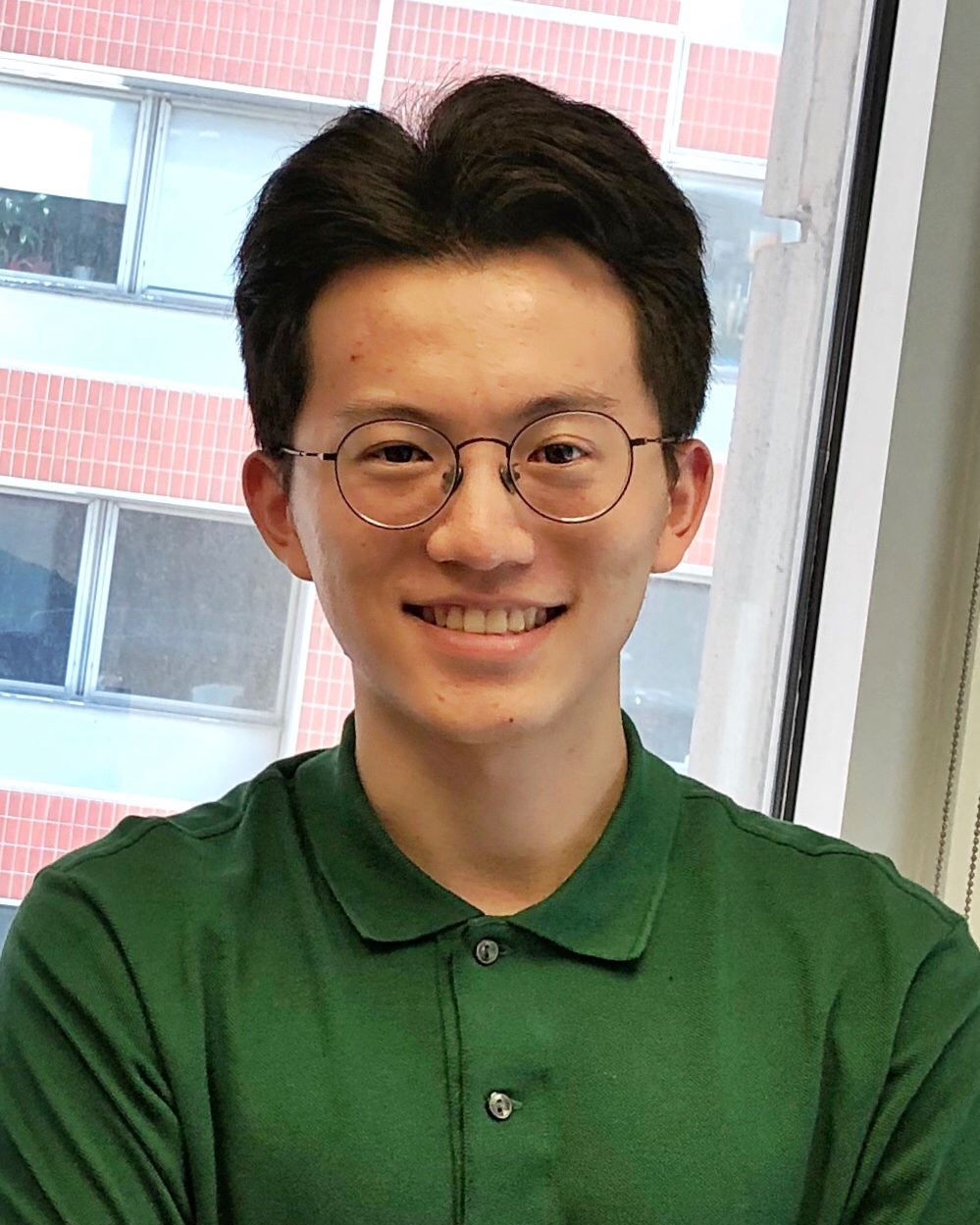}}]{Yulei Niu} is currently a Postdoctoral Research Scientist at the Fu Foundation School of Engineering and Applied Science, Columbia University. He received the Ph.D. and B.E. degrees in computer science from Renmin University of China, Beijing, China in 2020, and 2015, respectively. He was a Research Fellow in Nanyang Technological University, Singapore. His research interests include computer vision, causal inference, and machine learning.
\end{IEEEbiography}

\vspace{-4em}

\begin{IEEEbiography}[{\includegraphics[width=1in,height=1.25in,clip,keepaspectratio]{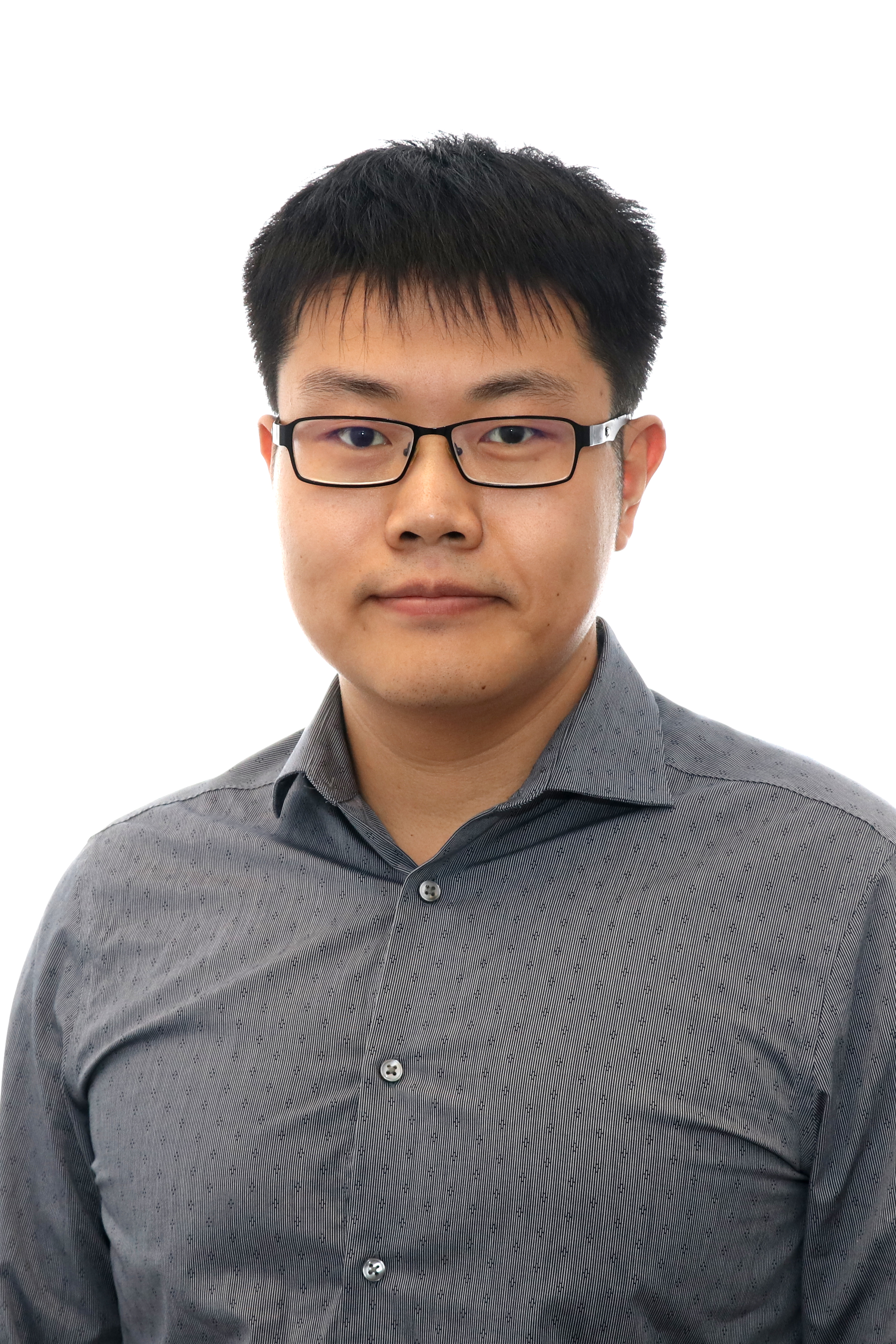}}]{Hanwang Zhang}
is currently an Assistant Professor at the School of Computer Science and Engineering, Nanyang Technological University, Singapore. He was a research scientist at the Department of Computer Science, Columbia University, USA. He has received the B.Eng degree in computer science from Zhejiang University, Hangzhou, China in 2009, and the Ph.D. degree in computer science from the National University of Singapore in 2014. 
\end{IEEEbiography}

\vspace{-4em}

\begin{IEEEbiography}[{\includegraphics[width=1in,height=1.25in,clip,keepaspectratio]{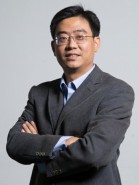}}]{Jun Xiao}
 received the Ph.D. degree in computer science and technology from the College of Computer Science, Zhejiang University, Hangzhou, China, in 2007. He is currently a professor with the College of Computer Science, Zhejiang University. His current research interests include computer animation, multimedia retrieval, and machine learning.
\end{IEEEbiography}




\end{document}